\documentclass[11pt]{article}
\usepackage[letterpaper,margin=0.8in]{geometry}
\usepackage[T1]{fontenc}
\usepackage{lmodern,microtype}
\usepackage{amsmath,amssymb,amsthm,graphicx,booktabs,array,tabularx,multirow,float,changepage}
\usepackage[numbers,sort&compress]{natbib}
\usepackage{algorithm,algpseudocode}
\usepackage[hidelinks]{hyperref}
\usepackage[nameinlink,noabbrev]{cleveref}
\graphicspath{{figures/}}
\newlength{\extralength}
\newcommand{\fulllength}{\textwidth}
\newtheorem{Theorem}{Theorem}
\newtheorem{Lemma}{Lemma}
\newtheorem{Proposition}{Proposition}

\newcommand{\authorcontributions}[1]{\section*{Author contributions}#1}
\newcommand{\funding}[1]{\section*{Funding}#1}
\newcommand{\institutionalreview}[1]{\section*{Ethics statement}#1}
\newcommand{\informedconsent}[1]{\section*{Informed consent}#1}
\newcommand{\dataavailability}[1]{\section*{Data and code availability}#1}
\newcommand{\acknowledgments}[1]{\section*{Acknowledgments and AI assistance}#1}
\newcommand{\conflictsofinterest}[1]{\section*{Competing interests}#1}
\title{Partial-Revelation Panel Forecasting: Predictive Information Gain, Allocation, and Coarsening-Consistent Operators}
\author{Bin Wen$^1$ \and Dai-Qiao Zhang$^2$ \and Tien-Ping Tan$^{1,*}$}
\date{}
\begin{document}
\maketitle
\begin{center}\small
$^1$School of Computer Sciences, Universiti Sains Malaysia, Gelugor 11800, Penang, Malaysia\\
$^2$School of Mathematical Sciences, Universiti Sains Malaysia, Gelugor 11800, Penang, Malaysia\\
$^*$Correspondence: \texttt{tienping@usm.my}
\end{center}
\begin{abstract}
Subjective labels are produced by panels of annotators, but annotation is often collected without measuring the information supplied by each additional response. We study partial-revelation panel forecasting: given the votes of \texorpdfstring{$k$}{k} of \texorpdfstring{$n$}{n} annotators on the previous dialogue turn, predict the full count distribution over the next turn's panel. We distinguish this cross-turn predictive information gain from the more general notion of same-item annotation value. For any nested revelation mechanism, the population decrement satisfies \texorpdfstring{$V(k)-V(k + 1)=I(Y;R_{k+1}\mid X_k)$}{V(k)-V(k+1) = I(Y; R(k+1) | X(k))}; coarsened-completely-at-random (CCAR) masking supplies the exchangeable random-rater interpretation used by our operator construction. For a fitted model, the decrement additionally contains the difference between adjacent excess predictive risks. We define coarsening-consistent panel operators and give a closed-form Dirichlet-multinomial realization, together with learned completion and rater-count transfer variants. Against per-level lookup, the cold-start count negative log-likelihood (NLL) differences are \texorpdfstring{$-$0.422/$-$0.166/$-$0.014}{-0.422/-0.166/-0.014} nats on IEMOCAP/M3ED/EmoWOZ. A new held-out rater experiment measures same-item value directly: the first and second additional votes reduce NLL by \texorpdfstring{$0.583/0.221$}{0.583/0.221} nats on IEMOCAP, \texorpdfstring{$0.578/0.184$}{0.578/0.184} on M3ED, and \texorpdfstring{$0.180/0.069$}{0.180/0.069} on EmoWOZ. Cross-turn and same-item gains are not interchangeable, and the CCAR masks are controlled interventions rather than observed arrival processes. Target-free allocation policies beat uniform only where row-level signal is predictable; at budget one they recover 83--93\% of retrospective feature class headroom on IEMOCAP/M3ED but do not improve EmoWOZ. The numerical gains are modest and corpus-dependent: beyond the cold-start result, several partial-revelation differences are small or null, especially on EmoWOZ. The implementation uses fixed seeds; code and aggregate result artifacts accompany this preprint.
\end{abstract}

\section{Introduction}
\label{sec:intro}

Datasets for subjective tasks---emotion, toxicity, stance---are built by
paying several annotators to label each item. How many should each item
get? Should every item get the same number? If annotators differ, which
one should a given item go to? These are ordinary budget questions asked
on every annotation project, and~they are usually answered by convention
(``three annotators per item'') rather than by~measurement.

This paper studies one such question as a forecasting problem. Consider a
system that must predict a future annotator panel while annotation of the
current material is still in progress: some annotators have submitted their
labels, others have not. The~available evidence is a \emph{partially
revealed} panel, and~the natural forecast target is the full count
distribution over the next item's panel rather than a single majority
label. We call this \emph{partial-revelation panel forecasting}. It defines
a sequential, cross-turn estimand: how much does an additional revealed
vote improve prediction of the next panel? This is related to, but~not the
same as, the~same-item annotation-budget question of predicting a remaining
vote for the item currently being annotated; we evaluate both estimands
separately.

\subsection{{The Cross-Turn Gain Has an Information-Theoretic Identity}}
Let $V(k)$ be the expected count negative log-likelihood of the target
panel given $k$ revealed annotators. For~the true conditional family,
$V(k)-V(k+1)=I(Y;R_{k+1}\mid X_k)$, so each increment of the cross-turn
revelation curve is a conditional mutual information. This is an identity
for the true conditional distribution, not an automatic claim about a
finite-sample fitted curve or about generic same-item annotation value. We
therefore add a direct held-out rater experiment and report the fitted-risk
correction explicitly in Section~\ref{sec:theory}.

If evaluator identities are observed and stable, the~same argument runs on
the subset lattice: writing $S$ for the set of annotators already revealed,
$V(S)-V(S\cup\{j\})=I(Y;R_j\mid X_S)$. This raises a routing question
that the exchangeable formulation cannot express. Our raw-data audit shows
that the original IEMOCAP parser did not provide stable columns on all
rows, so the revised manuscript treats identity-aware routing as a
sensitivity analysis rather than as a headline~result.

\subsection{Operators That Behave Sensibly Across Revelation Levels}
To estimate these curves one needs a forecaster defined at every $k$, and~a family of separately fitted models is not obviously coherent: a forecast
should not change its meaning merely because some votes are still hidden.
We formalize that requirement as \emph{coarsening consistency}: the
forecast at a partially revealed panel should equal the conditional
average of the full-panel forecast over its hidden completions. A~Dirichlet-multinomial completion model realizes this requirement in closed
form, as~a finite mixture over the enumerable completions of the hidden
votes. Because~the completion latent is small (at most 462 terms in our real-data
settings, including 56 when $n=3$ and $K=6$), every operator in the family admits exact marginalization rather
than sampling or relaxation; the formal definition and proof are given in
Section~\ref{sec:theory}.

\subsection{What We Found, Including What We Did Not Expect}
The axiom is natural and auditable, but~it is not what drives predictive
accuracy: a per-level marginalizer that violates it matches or beats the
consistent operator. Consistency is better understood as a structured
regularizer---costly where per-level cells are dense, beneficial where
they are sparse---and as a deployment property, since one table then
serves every $k$ and every panel size. Learning the completion helps, and~an information set control attributes that gain to context entering the
completion channel rather than to added capacity. We also test the simpler
explanation that the cross-turn signal is affective inertia by conditioning
on partial majority and speaker transition. Because~evaluator identifiers
are not consistently available across the corpora, we do not make a
general rater-routing~claim.

We also report three findings that contradict claims we ourselves made
earlier in this work: exact-marginal training of the completion head is
worse than the cross-entropy surrogate it was meant to replace; the
apparent superiority of a Dirichlet head over a cross-entropy head is largely
explained by the likelihood family rather than the training loss in this
protocol; and the
usual oracle used to bound allocation policies is clairvoyant, so the
commonly quoted ``fraction of oracle recovered'' understates what a
policy achieves. Section~\ref{sec:negative} collects~these.

\subsection{Contributions}
(i)~The partial-revelation forecasting object, its cross-turn
conditional mutual-information characterization, and~a direct same-item
held-out rater validation. (ii)~Coarsening-consistent panel operators with a
closed-form Dirichlet-multinomial realization, an~EM-inspired posterior-weighted moment estimator, a~rater-count-invariant extension, a~cell density shrinkage variant, and~a
learned completion operator with an information set control. (iii)~Ablation
and sensitivity evidence that probes the role of affective inertia and
speaker transition persistence, tests CCAR-simulated and non-CCAR masking, and~compares allocation
against uncertainty and margin heuristics. (iv)~A four-corpus evidence
chain with official held-out splits, cluster bootstraps, multiplicity-aware
reporting, and~deterministic numeric, logic, and~data quality audits. We
make no state-of-the-art claim; the contribution is a scoped measurement
framework and an auditable design~principle.

The remainder of the paper is organized as follows. Section~\ref{sec:related}
reviews disagreement-aware learning, repeated labeling, and~active
annotation. Section~\ref{sec:problem} defines the forecasting object and
the consistency axiom, Section~\ref{sec:theory} gives the population and
fitted-risk results, and~Section~\ref{sec:method} describes the operator
families. Section~\ref{sec:experiments} reports the real-corpus results,
including the new same-item and inertia controls; Sections~\ref{sec:negative}
and~\ref{sec:discussion} state corrections and practical implications.
Limitations and conclusions follow, with~implementation details and full
tables in the~Appendix.

\section{Related~Work}
\label{sec:related}

\subsection{Learning from Disagreement} 
Treating annotator disagreement as
signal is an established line with its own benchmark
series~\citep{lewidi2025}: soft-label prediction, perspectivist
evaluation and distribution-aware metrics~\citep{beyondconsensus2026}.
Representative modeling choices include evidential estimation of the
current utterance's distribution~\citep{wu2024ambiguity}, Dirichlet-prior
distributional emotion recognition in conversation (ERC)~\citep{wu2022disterc}, incomplete-label
LDL~\citep{li2024inldl,zhang2026simlabel}, and~analyses of hard versus
soft label targets~\citep{hardsoft2026}. All estimate the \emph{current}
item's distribution; none forecasts a future panel or conditions on a
partially revealed previous~one.

\subsection{How Many Annotators Are Worth It} 
Closest to our
revelation-value curve, \citet{saturation2026} subsample annotators on
ChaosNLI and show that different metrics saturate at different annotator
counts. That study measures metric-dependent saturation empirically; we characterize the increment exactly as a conditional
mutual information (Theorem~\ref{thm:cmi}), which additionally yields the
allocation consequences that a saturation curve alone does not, and~we
predict the curve per row rather than per~corpus.

\subsection{Repeated Labeling and Crowd Allocation} 
Repeated-labeling
strategies explicitly ask when a second or later label improves data quality
and how uncertainty can guide selective acquisition
\citep{sheng2008get,hsueh2009dataquality}. Active learning work for crowds
jointly considers which instance to query and which annotator to use
\citep{yan2011activecrowds,li2015multilabelcrowds}. More recent subjective
NLP work models annotator-specific heads or selects annotators to preserve
human label variation~\citep{wang2023actor,vandermeer2024acal}. These works
address pool-based acquisition or representation of disagreement. Our
setting is narrower: the item is already in a dialogue, a~partial panel is
revealed sequentially, and~the scoring target is the next item's full panel.
Accordingly, we include their uncertainty-style ideas as target-free
baselines, but~do not claim to reproduce their different acquisition
objectives.

\subsection{Emotion Forecasting; Multi-Rater and Coarsened Data}
Addressee-aware next-turn inference~\citep{li2021emotion},
listener emotion forecasting~\citep{koga2024forecasting},
Hi-EF~\citep{wang2026hief} and longitudinal affect~\citep{noor2026affect}
predict labels or events, not panel distributions.
BSETD~\citep{inoshita2026bsetd} builds Dirichlet-multinomial (DM) posteriors over soft-label
transition matrices for structure discovery, without~forecasts or proper
scores. Classical coarsened-data theory~\citep{dickey1987censored} covers
single multinomial observations with missing category distinctions, not
dialogue panels or consistency across revelation levels. To~our
knowledge, coarsening-consistent operators for future panel distributions
under partial revelation, and~the mutual-information characterization of
revelation increments, are~new.

\section{Partial-Revelation Panel~Forecasting}
\label{sec:problem}

\subsection{Notation} An utterance $u_t$ has an exchangeable count panel
$c_t \in \{0,\ldots,n\}^K$ with $\sum_j c_{t,j} = n$, where $K$ is the
number of emotion classes and $n$ the number of raters ($n = 3$ on
IEMOCAP, M3ED, and~EmoWOZ; $n = 5$ on EmotionLines). A~transcript-derived
condition $r_t$ (speaker same/switch, or~a bridging condition for
task-oriented corpora) accompanies the target. At~forecast time, a~subset
of $k \le n$ raters of the \emph{previous} panel is revealed under a
CCAR-simulated mask, yielding partial counts
$\tilde c^{(k)} \in \{0,\ldots,k\}^K$ with $\sum_j \tilde c^{(k)}_j = k$.
The forecasting object is the full conditional count distribution
\begin{equation}
Q_k(c_t \mid \tilde c^{(k)}, r_t),
\qquad k = 0,\ldots,n ,
\end{equation}
evaluated by count negative log-likelihood (NLL) against the full target panel $c_t$. Count NLL is a
strictly proper scoring rule for full-support distributions over count
vectors; coherence is assessed through the audited consistency gaps and
fitted-risk behavior through revelation-monotonicity checks (Sections~\ref{sec:ablation} and
\ref{sec:ratercount}).

\subsection{Worked Example} For $K = 2$, $n = 3$, and~$k = 2$, suppose the
revealed partial panel is $(1,1)$. The~hidden completion is either $(1,0)$
or $(0,1)$, so a consistent operator has the explicit form
$Q_2(\cdot\mid(1,1),r)=p\,Q_3(\cdot\mid(2,1),r)+(1-p)\,Q_3(\cdot\mid(1,2),r)$,
where $p$ is the completion posterior. This small case makes clear both
the finite marginalization and the meaning of a common full-panel
semantics; the general operator enumerates the analogous count
compositions.

\subsection{Coarsening Consistency} A family $\{Q_k\}_{k=0}^{n}$ is
\emph{coarsening-consistent} if for every partial panel
\begin{equation}
\label{eq:consistency}
Q_k(c_t \mid \tilde c^{(k)}, r) \;=\;
\mathbb{E}_{c' \sim P(\cdot \mid \tilde c^{(k)}, r)}\,
Q_n(c_t \mid c', r),
\end{equation}
where the completion posterior $P(c' \mid \tilde c^{(k)}, r)$ is taken from
the model itself. In~words: the operator for a partially revealed panel is
the completion-marginalized operator for the full panel. Per-level lookup
tables---a separate table re-fit at each $k$, indexed by partial composition
and relation---need not satisfy the axiom; departures are measured by the
Kullback--Leibler (KL) divergence:
\begin{equation}
\mathrm{Gap}(k) \;=\;
\mathrm{KL}\Bigl( Q_k^{\mathrm{lookup}}(\cdot\mid\tilde c^{(k)},r)
\;\Big\|\;
\mathbb{E}_{c'}\bigl[Q_n^{\mathrm{lookup}}(\cdot\mid c',r)\bigr]
\Bigr).
\label{eq:gap}
\end{equation}
At
 $k = 0$ the lookup table has a single partial-composition key and
therefore degenerates to the relation-conditioned marginal panel
distribution: it discards all transition structure. This degeneracy is the
confound that motivates our mechanism ablation (Section~\ref{sec:ablation}).

\subsection{Estimands and Masking Interpretation} The primary cross-turn
increment is $\Delta_k^{\mathrm{cross}}=V_q(k)-V_q(k+1)$, where the target
is the next turn's full panel. A~same-item increment instead holds an item
fixed, designates one annotator as the target, and~measures the NLL
reduction from revealing additional annotations among the other annotators.
These quantities answer different operational questions and are compared
rather than conflated in Section~\ref{sec:sameitem}. The~corpora do not
provide a complete timestamped record of annotation arrival. Thus CCAR and
the identity- or label-dependent alternatives below are controlled masking
interventions; they are not claims about the historical arrival~mechanism.

\section{Population and Fitted Predictive~Information}
\label{sec:theory}

Let $X_k$ denote the information available after $k$ reveals of the previous
panel. We assume that the revelation process is nested:
$\sigma(X_k)\subseteq\sigma(X_{k+1})$ and
$X_{k+1}=(X_k,R_{k+1})$, where $R_{k+1}$ contains the newly observed vote
and, when available, its rater identity or selection state. Let $Y$ be the
target panel, $P_k(\cdot) = P(Y \mid X_k)$ the true conditional family, and~$V(k) = \mathbb{E}[-\log P(Y \mid X_k)]$ the expected count~NLL.

\begin{Theorem}[Revelation monotonicity]
\label{thm:monotone}
For any strictly proper scoring rule $S$,
\[
\mathbb{E}[S(P_{k+1}, Y)] \;\le\; \mathbb{E}[S(P_k, Y)] .
\]
\end{Theorem}

The proof (Appendix~\ref{app:proofs}) is the standard Blackwell-style
argument: nestedness and the tower property give
$P_k=\mathbb{E}[P_{k+1}\mid X_k]$, and~strict propriety of $S$ then makes
the coarser forecast no better than the finer one. CCAR is not needed for
this information-ordering result. In~this paper, CCAR is the controlled,
exchangeable revelation design used for the Dirichlet-multinomial completion
posterior, the~random-rater interpretation, and~the real-corpus benchmark;
the identity- and label-dependent mechanisms are sensitivity interventions.
The useful reading is diagnostic rather than predictive: \emph{violations}
of population monotonicity by a fitted curve diagnose differences in
adjacent excess predictive risk through finite-sample estimation,
approximation, or~model misspecification, not a failure of the population
information~identity.

\begin{Theorem}[Revelation increments are conditional mutual information]
\label{thm:cmi}
Let $R_{k+1}$ be the newly revealed information at step $k + 1$ under any
nested revelation mechanism as above. Then
\begin{equation}
\label{eq:cmi}
V(k) - V(k+1) \;=\; I(Y;\, R_{k+1} \mid X_k),
\end{equation}
and consequently $V(0) - V(n) = I(Y; X_n\mid X_0)$.
\end{Theorem}

\begin{proof}[Proof sketch]
$V(k)-V(k+1) = \mathbb{E}[\log P(Y\mid X_{k+1}) - \log P(Y\mid X_k)]$,
which is $\mathbb{E}_{X_k,R_{k+1}}\bigl[\mathrm{KL}(P(Y\mid X_k,R_{k+1}) \;\|\;
P(Y\mid X_k))\bigr] = I(Y;R_{k+1}\mid X_k)$ by the standard decomposition
of conditional mutual information into expected KL.
\end{proof}

For a fitted conditional model $q_k$ evaluated under the data-generating
distribution, define
\begin{equation}
V_q(k)=\mathbb{E}\left[-\log q_k(Y\mid X_k)\right],
\qquad
\varepsilon_k=\mathbb{E}_{X_k}
\mathrm{KL}\left(P(Y\mid X_k)\,\|\,q_k(Y\mid X_k)\right).
\end{equation}
The fitted curve is therefore not itself a population conditional mutual information (CMI) estimator unless
the adjacent conditional models have equal excess~risk.

\begin{Proposition}[Fitted-risk decomposition]
\label{prop:fitted-risk}
For any full-support fitted family $\{q_k\}_{k=0}^{n}$,
\begin{equation}
V_q(k)-V_q(k+1)
=I(Y;R_{k+1}\mid X_k)+\varepsilon_k-\varepsilon_{k+1}.
\label{eq:fitted-risk}
\end{equation}
Consequently, finite-sample fitted curves need not be monotone or satisfy
diminishing marginal returns, even though the corresponding Bayes curve is
monotone.
\end{Proposition}

\begin{proof}[Proof sketch]
The logarithmic-score identity gives
$\mathbb{E}[-\log q_k(Y\mid X_k)]=H(Y\mid X_k)+\varepsilon_k$.
Subtracting the corresponding expression at $k+1$ and applying
Theorem~\ref{thm:cmi} yields~\eqref{eq:fitted-risk}.
\end{proof}

\begin{figure}[H]
\centering
\includegraphics[width=.999\textwidth]{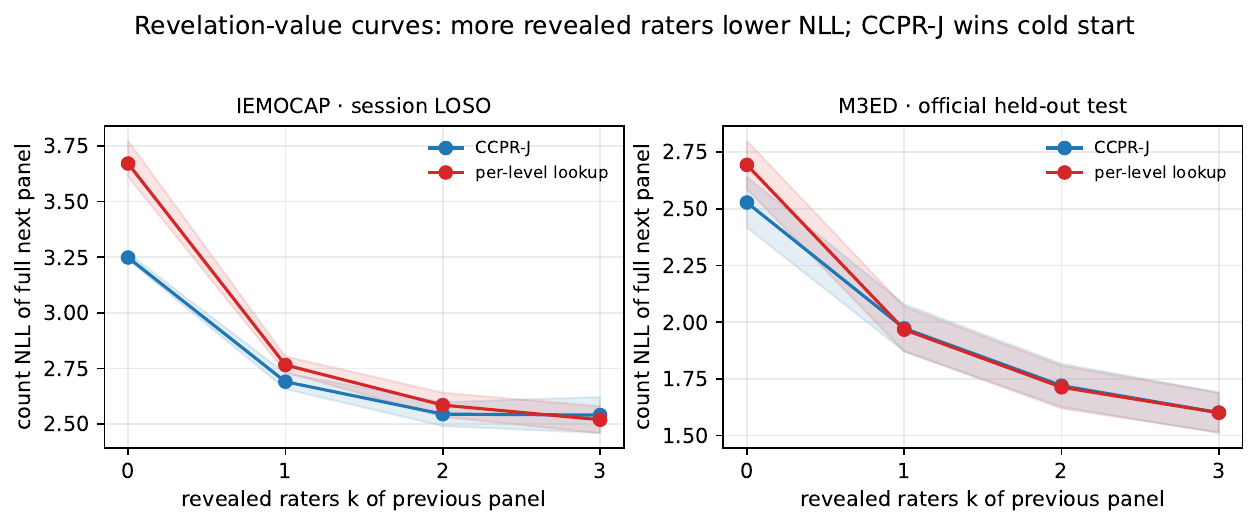}
\caption{Fitted revelation curves (expected count NLL) for ccprJ and
per-level lookup on IEMOCAP (session-level leave-one-session-out (LOSO)) and M3ED (official held-out).
The shaded bands are 95\% cluster-bootstrap confidence intervals (session clusters for IEMOCAP and dialogue clusters for M3ED).
Theorem~\ref{thm:cmi} applies to the population curve; deviations of a
fitted curve from monotonicity are fitted-risk diagnostics described by
Proposition~\ref{prop:fitted-risk}.}
\label{fig:curve}
\end{figure}

The fitted curves are shown in Figure~\ref{fig:curve}.

Two consequences follow. First, the~Bayes increments are non-negative, but~a fitted decrement is a model-based diagnostic whose error is governed by
the adjacent excess risks in Proposition~\ref{prop:fitted-risk}. Second,
greedy water-filling is optimal when the \emph{expected marginal gains}
$D_i(k)=V_i(k)-V_i(k + 1)$ are non-increasing in $k$. Equivalently, the~cost curve has non-negative second differences
$V_i(k)-2V_i(k + 1)+V_i(k + 2)\geq0$; this is diminishing marginal return
for a decreasing loss, or~concavity of the utility $-V_i$, rather than
standard concavity of the loss itself. This is an assumption for the
allocation result, not a property guaranteed by finite-sample fitted curves.
Section~\ref{sec:budget} tests the resulting policies directly on held-out
rows and reports the empirical diminishing-return~rate.

\begin{Proposition}[Greedy water-filling under diminishing marginal returns]
\label{prop:greedy}
If each row's expected marginal gains
$D_i(k)=V_i(k)-V_i(k + 1)$ are non-increasing in $k$, then for any integer
budget $B$, an~allocation that repeatedly awards the next rater to the row
with the largest current marginal gain
$V_i(k_i)-V_i(k_i + 1)$ minimizes $\sum_i V_i(k_i)$ subject to
$\sum_i k_i = B$ \citep{fox1966marginal}.
\end{Proposition}

The proof is the classical marginal-analysis argument of
\citet{fox1966marginal}: non-increasing marginal gains make the sorted list
of all gains a valid optimizer, and~the greedy exchange step preserves
optimality. The~assumption is about \emph{expected} curves; individual rows
are noisy (Section~\ref{sec:budget} reports the empirical
diminishing-return rate), which is precisely why the policy is calibrated on
bin-averaged curves rather than single~rows.

\section{Coarsening-Consistent Operator~Families}
\label{sec:method}

\subsection{Response Table and DM Closure} The full-panel operator is a
Dirichlet-multinomial response table
\begin{equation}
Q_n(c_t \mid c', r) \;=\;
\mathrm{DM}\bigl(c_t \mid \kappa\, T_{g(c', r)}\bigr),
\end{equation}
where $g(\cdot)$ maps the previous full composition and relation to a row
of $T$, $\kappa$ is a concentration, and~$T$ is a distribution over the $K$
classes. Assume a shared label-probability vector $\theta\sim\mathrm{Dirichlet}(\beta)$,
with $\beta_j>0$, conditionally independent categorical votes given $\theta$,
and an ignorable CCAR revelation subset. The completion
posterior for the $n-k$ unobserved votes is
$\mathrm{DM}_{n-k}(m \mid \beta + \tilde c^{(k)})$ by conjugacy and rater
exchangeability (Lemma~\ref{lem:dm}); hence the consistent operator is a
finite~mixture
\begin{equation}
\label{eq:consistent-op}
Q_k(c_t \mid \tilde c^{(k)}, r) \;=\;
\sum_{m:\,|m|=n-k}
\mathrm{DM}\bigl(c_t \mid \kappa\, T_{g(\tilde c^{(k)}+m, r)}\bigr)\,
\mathrm{DM}_{n-k}\bigl(m \mid \beta + \tilde c^{(k)}\bigr),
\end{equation}
with at most $\binom{n-k+K-1}{K-1}$ terms (56/21/6/1 for
$k = 0/1/2/3$, $K = 6$). Mixtures of consistent families sharing the same
completion posterior are consistent by linearity of expectation
(Lemma~\ref{lem:mixture}) when a single mixing weight is used across
levels. The per-level development-selected neural blends below remain valid
probability distributions but are not claimed to inherit exact coarsening
consistency.

\subsection{CCPR-J (Exact Consistency)} The shared response table $T$ is
estimated jointly across all revelation levels by an EM-inspired
posterior-weighted iteration (Algorithm~\ref{alg:em}). The responsibility step computes, for~each training row and
each completion $m$ of its revealed partial panel, the~posterior weight
under the current $T$; the moment step re-estimates each row of $T$ as a
smoothed, weighted mean of target votes. This moment update is not an exact
maximizer of the fixed-concentration Dirichlet--multinomial likelihood;
likelihood monotonicity and convergence to an EM optimum are not guaranteed. Per-level empirical completion
tables $W_k(m \mid \tilde c^{(k)}, r)$ are fit on train-only masked data;
the forecast is \eqref{eq:consistent-op} with $W_k$ in place of the
conjugate posterior. The iteration weights revelation levels
$(1,1,1,4)$ to balance full-revelation fidelity; Section~\ref{sec:robust}
shows results are insensitive to this~choice.

\subsection{CCPR-D (Deployment Variant)} Exact consistency has a small
cost at $k=n$: the jointly estimated table differs slightly from the
dedicated full-panel table. CCPR-D therefore uses the dedicated full-panel
table at $k=n$ and the shared completion machinery elsewhere; consistency
becomes approximate, and~the audited gap ($0.003$--$0.008$ on IEMOCAP,
vs.\ $\approx 10^{-17}$ for CCPR-J and $0.02$--$0.05$ for per-level lookup)
is reported rather than~hidden.

\subsection{CCPR-N (Rater-Count Invariance)} The per-vote formulation makes
the operator rater-count-agnostic: $T$ is a per-vote distribution over an
$n$-independent summary of the previous panel---majority label, capped
support, and~relation---and the target panel is $\mathrm{DM}(n, \kappa T_s)$
for any $n$. The~summary key is deliberately coarser than the full
composition key; this trade-off is measured directly by the $n = 3$ rows of
Table~\ref{tab:n5}. It enables zero-shot rater-count transfer
($n = 3\to n = 5$) and mixed-$n$ deployment with a single~table.

\subsection{CCPR-S (Cell Density Shrinkage)} The bias-variance analysis of
Section~\ref{sec:scan} suggests a direct algorithmic realization: instead
of committing to either the shared consistent table or a per-level refit,
interpolate in probability space with a weight controlled by cell density,
\begin{equation}
Q^{\mathrm{CCPR-S}}_{k} \;=\;
\bigl(1-\lambda_k\bigr)\, Q^{\mathrm{shared}}_{k}
 \;+\; \lambda_k\, Q^{\mathrm{per-level}}_{k},
\qquad
\lambda_k \;=\; \frac{\bar c_k}{\bar c_k + h_k},
\end{equation}
where $\bar c_k$ is the average number of training rows per per-level cell
at level $k$ and $h_k$ is one bandwidth per level selected on the inner dev
split. Dense cells ($\bar c_k \gg h_k$) use per-level flexibility
($\lambda_k\to 1$); sparse cells shrink toward the shared table
($\lambda_k\to 0$). With one common completion kernel and one mixing
weight held constant across all levels, joint convexity of KL bounds the
mixture's consistency gap (Proposition~\ref{prop:gapmix}). The level-specific
weights used by CCPR-S need not satisfy these conditions, so no corresponding
analytic gap guarantee is claimed for this implementation. CCPR-S is a
heuristic, development-selected relaxation of the exact operator~family.

\section{Results}
\label{sec:experiments}

\subsection{Protocol}
IEMOCAP~\citep{busso2008iemocap} uses
leave-one-session-out over five sessions. The~reported 4{,}883 strict rows
are the rows for which both the target and immediately preceding utterance
have exactly three single-label categorical judgments from the six standard
classes; this is a subset of 7{,}741 strict target utterances in the raw
categorical files ($K = 6$, $n = 3$; cluster bootstrap over sessions).
M3ED~\citep{zhao2022m3ed} uses
5-fold series-blocked development and one final run on the official
11-series held-out test ($K = 7$, $n = 3$; selection only on the official
7-series validation split; dialogue bootstrap). EmoWOZ
DialMAGE~\citep{feng2022emowoz} uses the official 796/100/100 dialogue
split ($K = 7$, $n = 3$; dialogue bootstrap). EmotionLines Friends~\citep{chen2018emotionlines}
($n = 5$, $K = 7$) uses 5-fold dialogue CV (13{,}503 rows; dialogue
bootstrap) plus an external eval-gold set. For the cross-turn runs, $k \in
\{0,\ldots,n\}$ raters of each previous panel are revealed under a fixed
seed; in the sensitivity and inertia experiments the masks are nested
prefixes of one per-row order; all smoothing, concentration, and~iteration/mixture
hyperparameters are
selected on inner dev splits; held-out units never enter selection. IEMOCAP
and M3ED use text+visual context in the main multimodal runs, with~matched
text-only sensitivity runs; EmoWOZ uses train-only term frequency--inverse
document frequency (TF--IDF) context because
the released DialMAGE representation has no aligned visual stream. The implementation uses fixed seeds. We report separate
comparison families for the nine learned completion contrasts, the same-item
increments, and the inertia-control contrasts. Holm correction is used for
the primary nine-cell mechanism family. Benjamini--Hochberg adjustments for
the validation analyses are descriptive; the stored same-item q-values were
computed within each corpus's two increments, rather than over one
cross-corpus six-increment family. The~validation p-values are
uncentered bootstrap tail fractions (p = 0 means no bootstrap draw crossed
zero), so 95\% confidence intervals (CIs) and effect sizes are primary. Allocation and descriptive tables
are accompanied by block-bootstrap intervals and are not interpreted as a
collection of independent~discoveries.

\subsection{Main Results vs.\ Per-Level~Lookup}
\label{sec:main}

Table~\ref{tab:main} reports ccprJ minus per-level lookup by revelation
level. Cold-start forecasting ($k = 0$) wins on all three corpora with
cluster bootstrap CIs excluding zero (IEMOCAP $-0.422$ $[-0.505,-0.373]$,
M3ED $-0.166$ $[-0.206,-0.125]$, EmoWOZ $-0.014$ $[-0.023,-0.003]$).
Partial-revelation results are corpus-dependent: IEMOCAP wins $k = 1/2$
significantly; EmoWOZ wins $k = 1/2$ significantly and $k = 3$ within
$0.003$ nats (CI includes zero); M3ED loses by at most $0.007$ nats at
$k = 1/2/3$, none significant. At~$k = 3$ on IEMOCAP, ccprJ is $+0.021$
nats worse with a CI whose lower endpoint is $0.0002$ (four of five session-level differences
positive); the CCPR-D deployment variant removes this cost by construction
and ties the dedicated lookup table. We report this cell exactly and do
not claim all-$k$ wins. The~$k = 0$ degeneration is visible in the numbers
themselves: per-level lookup is within $0.03$ nats of the flat marginal on
every corpus (IEMOCAP $3.6703$ vs.\ $3.6836$; M3ED $2.6925$ vs.\ $2.7096$;
EmoWOZ $2.0565$ vs.\ $2.0850$).

For practical calibration, count NLL differences are log-score units per
target panel, not percentage-point accuracy changes. A~reduction of $d$
nats multiplies the geometric probability assigned to the observed panel
by $e^d$; thus the cold-start reductions of $0.422$, $0.166$, and~$0.014$
nats correspond to approximately $52.5\%$, $18.0\%$, and~$1.4\%$ higher
geometric score on IEMOCAP, M3ED, and~EmoWOZ, respectively. The~partial-level
differences are generally much smaller, so they should not be read as large
accuracy~gains.

\begin{table}[H]
\centering
\caption{Count NLL difference ccprJ \texorpdfstring{$-$}{-} lookup (nats), by~corpus and
revelation level. Negative favors ccprJ. Bold: 95\% cluster bootstrap CI
excludes zero in ccprJ's favor. Full NLL values in
Appendix~\ref{app:tables}.}
\label{tab:main}

\begin{adjustwidth}{-\extralength}{0cm}
\newcolumntype{C}{>{\centering\arraybackslash}X}
\newcolumntype{B}{>{\raggedright\arraybackslash}X}
\newcolumntype{A}{>{\raggedleft\arraybackslash}X}
\begin{tabularx}{\fulllength}{BCCCC}

\toprule

\textbf{Corpus} & \boldmath{$k = 0$} & \boldmath{$k = 1$} & \boldmath{$k = 2$} & \boldmath{$k = 3$} \\
\midrule
IEMOCAP & \boldmath{$-0.422$} & \boldmath{$-0.075$} & \boldmath{$-0.041$} & $+0.021$ \\
 & $[-0.505,-0.373]$ & $[-0.083,-0.067]$ & $[-0.049,-0.032]$ & $[0.000,0.046]$ \\
M3ED official & \boldmath{$-0.166$} & $+0.006$ & $+0.007$ & $+0.001$ \\
 & $[-0.206,-0.125]$ & $[-0.005,0.017]$ & $[-0.000,0.014]$ & $[-0.011,0.012]$ \\
EmoWOZ official & \boldmath{$-0.014$} & \boldmath{$-0.006$} & \boldmath{$-0.006$} & $-0.002$ \\
 & $[-0.023,-0.003]$ & $[-0.011,-0.001]$ & $[-0.010,-0.001]$ & $[-0.010,0.006]$ \\
\bottomrule

\end{tabularx}
\end{adjustwidth}

\end{table}

Negative controls confirm the operator uses the previous panel: permuting
its alignment costs $\approx 1.02/0.81/0.13$ nats at $k = 2$ on
IEMOCAP/M3ED/EmoWOZ, while permuting only the relation array costs
$\approx$0.08--$0.18$ nats on IEMOCAP/M3ED and $\approx$0.004 on~EmoWOZ.

\subsection{Same-Item Held-out Rater~Validation}
\label{sec:sameitem}

The cross-turn target is not automatically a same-item annotation value.
The same-item and inertia validations below use separate block
cross-validation protocols: IEMOCAP uses five-session LOSO (4883 rows);
M3ED uses five-fold series CV on its 45 official training and validation
series (9171 rows), excluding the official test series; EmoWOZ uses
five-fold dialogue CV on the 995 retained dialogues (11098 rows).
These validation results are distinct from the official-test main results.
We designate each rater in turn as a held-out target, reveal every
$k$-subset of the remaining raters, and~fit
$P(R_{\mathrm{held\mbox{-}out}}\mid\tilde c^{(k)})$ using only the outer
training blocks. The~procedure rotates the held-out rater and averages over
all subsets, so it does not assign a stable meaning to an annotation column
position. Table~\ref{tab:sameitem} reports the NLL reduction from the first
and second additional annotations. All intervals are block bootstraps over
the same units as the corresponding corpus~split.

\begin{table}[H]
\centering
\caption{Same-item held-out rater NLL reductions (nats). Each cell is the
reduction from \texorpdfstring{$k$}{k} to \texorpdfstring{$k + 1$}{k+1} revealed votes for a held-out annotator; 95\%
block-bootstrap CIs are in brackets.}
\label{tab:sameitem}
\newcolumntype{C}{>{\centering\arraybackslash}X}
\newcolumntype{B}{>{\raggedright\arraybackslash}X}
\newcolumntype{A}{>{\raggedleft\arraybackslash}X}
\begin{tabularx}{\textwidth}{BCC}

\toprule

\textbf{Corpus} & \boldmath{$0\to1$} & \boldmath{$1\to2$} \\
\midrule
IEMOCAP & $0.583$ $[0.565,0.600]$ & $0.221$ $[0.197,0.243]$ \\
M3ED & $0.578$ $[0.546,0.612]$ & $0.184$ $[0.174,0.195]$ \\
EmoWOZ & $0.180$ $[0.167,0.193]$ & $0.069$ $[0.066,0.072]$ \\
\bottomrule

\end{tabularx}

\end{table}

The direct experiment establishes a same-item value signal, but~it does not
validate the cross-turn proxy as an interchangeable estimand. On~IEMOCAP,
the row-wise Spearman correlation between same-item NLL and the existing
cross-turn ccprJ NLL is $\rho=0.266$ at $k = 0$ and $\rho=0.011$ at $k = 1$.
We therefore use the cross-turn curve only for sequential next-panel
forecasting and use the same-item curve when discussing repeated~labeling.

\subsection{Affective Inertia and Speaker Transition~Controls}
\label{sec:inertia}

To test whether the cross-turn signal is only emotional persistence, we fit
train-only smoothed target-panel tables under four information sets: no
previous-panel state (flat), speaker relation only, partial-majority plus
speaker relation, and~full partial composition plus speaker relation. A~single per-row random order is used to generate nested partial panels for all
four models and all adjacent levels. Table~\ref{tab:inertia} reports the NLL
reductions between adjacent revelation~levels.

\begin{table}[H]
\centering
\caption{Inertia controls: adjacent NLL reductions (nats) for majority and
full partial-composition controls, with~95\% block-bootstrap CIs. Ties in
partial majority are retained as ties. The~first step is shared because one
revealed vote determines the partial majority.}
\label{tab:inertia}

\begin{adjustwidth}{-\extralength}{0cm}
\newcolumntype{C}{>{\centering\arraybackslash}X}
\newcolumntype{B}{>{\raggedright\arraybackslash}X}
\newcolumntype{A}{>{\raggedleft\arraybackslash}X}
\begin{tabularx}{\fulllength}{llCCc}

\toprule

\textbf{Corpus} & \textbf{Control} & \boldmath{$0\to1$} & \boldmath{$1\to2$} & \boldmath{$2\to3$} \\
\midrule
IEMOCAP & partial majority + relation & $0.545$ $[0.498,0.591]$ & $0.116$ $[0.103,0.131]$ & $-0.042$ $[-0.077,-0.018]$ \\
 & full composition + relation & $0.545$ $[0.498,0.591]$ & $0.116$ $[0.103,0.131]$ & $0.048$ $[0.009,0.089]$ \\
M3ED & partial majority + relation & $0.640$ $[0.597,0.686]$ & $0.272$ $[0.236,0.308]$ & $-0.172$ $[-0.202,-0.142]$ \\
 & full composition + relation & $0.640$ $[0.597,0.686]$ & $0.272$ $[0.236,0.308]$ & $0.279$ $[0.249,0.309]$ \\
EmoWOZ & partial majority + relation & $0.057$ $[0.047,0.067]$ & $0.025$ $[0.018,0.032]$ & $-0.016$ $[-0.024,-0.008]$ \\
 & full composition + relation & $0.057$ $[0.047,0.067]$ & $0.025$ $[0.018,0.032]$ & $0.005$ $[-0.001,0.012]$ \\
\bottomrule

\end{tabularx}
\end{adjustwidth}

\end{table}

The first-revelation gain is largely explained by current-turn affective
state and speaker transition. The~full composition control retains later
gains on M3ED ($0.272$ and $0.279$ nats), on~IEMOCAP ($0.116$ and
$0.048$), and~a smaller first later gain on EmoWOZ ($0.025$; the final
$0.005$ interval crosses zero). In~contrast, the~majority-only control
shows a negative final increment on all three corpora, demonstrating that
majority-only compression can discard useful composition information even
when ties are retained. The negative fitted increments may reflect
approximation and estimation error; they do not imply negative population
information. This result narrows the mechanistic claim: the operator
forecasts a better estimate of the current affective state and its
continuation; it does not show that revealed votes are independent
annotation-value signals after all temporal structure is~removed.

\subsection{Synthetic Validation of Fitted-Risk and Masking~Sensitivity}
\label{sec:synthetic}

We next generated panel data from a known conditional distribution so that
the population increments could be enumerated exactly. We varied
$n\in\{3,5,10\}$, the~strength of previous-panel dependence
$\rho\in\{0.1,0.5,0.9\}$, training size, and~five independent random
replicates. Every simulated path is nested across $k$. For~identity slot
masking, labels are exchangeably assigned to slots before slots with
different arrival weights are sampled; consequently its population
count composition law is the same as CCAR. The~label-dependent mechanism
samples remaining labels sequentially with unequal label weights. The~population references for all three mechanisms are computed by exact
composition enumeration, while the empirical decrement uses a train-only
smoothed composition table. Table~\ref{tab:synth} summarizes the central
$\rho=0.5$ settings; the full grid and seeds are in 
the reproducibility~package.

\begin{table}[H]
\centering
\caption{Synthetic sensitivity at \texorpdfstring{$\rho=0.5$}{rho = 0.5} (five replicates). Bias and
RMSE are ranges/maxima across adjacent increments; violation is the fraction
of replicates whose fitted curve violates monotonicity at any level.}
\label{tab:synth}
\newcolumntype{C}{>{\centering\arraybackslash}X}
\newcolumntype{B}{>{\raggedright\arraybackslash}X}
\newcolumntype{A}{>{\raggedleft\arraybackslash}X}
\begin{tabularx}{\textwidth}{lcCcCC}

\toprule

\textbf{Mechanism} & \boldmath{$n$} & \textbf{Train Size} & \textbf{Bias Range} & \textbf{Max RMSE} & \textbf{Violation} \\
\midrule
CCAR & 3 & 300 & $[-0.137,-0.086]$ & 0.147 & 0.80 \\
CCAR & 3 & 1000 & $[-0.081,-0.036]$ & 0.083 & 0.20 \\
CCAR & 3 & 3000 & $[-0.030,-0.018]$ & 0.032 & 0.00 \\
CCAR & 5 & 1000 & $[-0.115,-0.078]$ & 0.118 & 1.00 \\
identity slot & 3 & 1000 & $[-0.074,-0.042]$ & 0.076 & 0.00 \\
label-dependent & 5 & 1000 & $[-0.118,-0.087]$ & 0.118 & 1.00 \\
\bottomrule

\end{tabularx}

\end{table}

Under CCAR with $n = 3$, increasing the training size reduces both bias and
fitted-curve monotonicity violations, while the $n = 5$ high-dimensional
composition table remains unstable at $1000$ rows. This reproduces the
interpretation of the observed EmotionLines reversal: $V(4)<V(5)$ is a
finite-sample fitted-risk effect, not a contradiction of the population
theorem. The~fitted-risk decomposition in
Proposition~\ref{prop:fitted-risk} holds to numerical precision (maximum
absolute residual $0.0$ in every setting), but~this zero is an algebraic
check because the excess-risk term is defined from the two reported curves.
The exact population reference is monotone in all 39 settings, including
identity slot and label-dependent masking. Thus, the finite fitted violations
under those mechanisms should not be described as failures of the population
theorem; they are sensitivity evidence about estimator error under
non-CCAR masking, not measurements of the corpora's unobserved arrival
process.

\subsection{Mechanism Ablation: The Cost of Response-Table Sharing}
\label{sec:ablation}

To separate marginalization from consistency, we fit per-level response
tables $T_k$ by the same posterior-weighted moment procedure restricted to level-$k$ rows, with~the same completion tables $W_k$, concentration, and~smoothing as CCPR-J.
This \emph{inconsistent marginalizer} has identical information access---it
can complete the unobserved previous panel at $k = 0$---but 
violates~\eqref{eq:consistency}. A~shared variant with equal $k$-weights
(ccprJ-eq) removes the confounding $(1,1,1,4)$ weight. The~complete
ccprJ-eq versus inconsistent comparison, including confidence intervals,
is reported in Appendix~\ref{app:operator_tables}; positive differences
mean that the shared table is~worse.

In the full-data $n=3$ ablation, consistency gives no significant
predictive improvement; on two corpora it costs up to $0.048$~nats at low
$k$, while differences on EmoWOZ are small. The~inconsistent marginalizer also beats per-level lookup at every
$k$ on M3ED and at $k = 0$ on IEMOCAP ($-0.494$). At~$n = 5$ the pattern
repeats ($+0.032$/$+0.023$/$+0.014$/$+0.004$ at $k = 0..3$, $-0.006$ at $k = 4$,
$0$ at $k = 5$). The~cost concentrates at partial revelation, where
per-level refits have the most freedom. These results establish consistency as a coherence property, rather
than a general predictive advantage; sparse cells can nevertheless benefit
from sharing (Section~\ref{sec:scan}). Operationally, one table
serves every $k$ and, via CCPR-N, every $n$; the axiom is auditable
($\approx$$10^{-17}$ gap for CCPR-J, $0.02$--$0.05$ for lookup).

\subsection{Learned Completion: CCPR-L}
\label{sec:learned}

Table~\ref{tab:abl} holds the empirical completion distribution fixed
and compares shared versus per-level response-table estimation, isolating
the cost of sharing. We now change the completion distribution to test
whether context can improve prediction of the unobserved previous-panel
votes. The~empirical
completion table conditions only on the partial composition and the
speaker relation and cannot use the context window that the neural
mixture (Section~\ref{sec:neural}) already exploits for the target.
CCPR-L learns the completion: per level $k$, a~GRU over the context
window, the~partial-panel broadcast, the~relation, and~the empirical
prior's log-probabilities predicts the missing $n{-}k$ votes as a softmax
over the level's completion compositions (cross-entropy on the true
completion, three seeds averaged). The~response table is refit by the same
posterior-weighted moment iteration but with these row-specific weights, and~a dev-selected per-level
scalar shrinks the learned completion toward the empirical one,
$W_k^\star = (1-\lambda_k)W_k^{\mathrm{emp}} + \lambda_k
W_k^{\mathrm{learned}}$ with $\lambda_k \in \{0,.25,.5,.75,1\}$. At~$\lambda_k = 0$ CCPR-L reduces exactly to the inconsistent baseline, so
$\lambda_k$ is a calibrated fallback when context does not transfer
(M3ED $k = 2$); it is one scalar per level, not $(n + 1)$ tables.

The matched comparison is CCPR-L$+$head versus ccprJ$+$head: same target
head, same dev-selected blending, same response vocabulary; only the
completion differs. It is nevertheless a \emph{composite}, since CCPR-L
uses per-level response tables and so inherits the per-level-vs-joint gap
of Section~\ref{sec:ablation}. To~attribute the gain we add an
information set control, CCPR-L$_{\mathrm{noctx}}$: the identical model
with the completion head's context window zeroed, holding architecture,
seeds, grids, dev rows, target head, input dimensionality and parameter
count fixed (the state block is concatenated, not overwritten), so it
removes context and nothing else. Appendix~\ref{app:ccprl} reports the
matched gain and splits it into \emph{context} (CCPR-L $-$ CCPR-L$_{\mathrm{noctx}}$)
and the residual table gap. The~context term is significant in 8 of the 9
partial-revelation cells (all but M3ED $k = 2$) and dominates at $k = 0$
(93/78/85\% of the total); the residual reproduces
Section~\ref{sec:ablation} and accounts entirely for the $k = 3$ losses,
which we do not claim as a CCPR-L effect. Holm correction over these nine
cells rejects the same eight at $\alpha = 0.05$ (bootstrap $p\le0.0246$,
EmoWOZ $k = 2$ marginal; M3ED $k = 2$ retains $H_0$ at $p = 0.27$), so
the claim does not rest on ignoring~multiplicity.

Three signatures support the mechanism. (i)~The context term decreases
strictly in $k$ on all three corpora ($0.061\!>\!0.040\!>\!0.015$;
$0.074\!>\!0.016\!>\!0.001$; $0.015\!>\!0.007\!>\!0.004$) and is exactly
$0.000$ at $k = n$, where the completion is empty---a structural zero the
mechanism predicts rather than fits. (ii)~Dev selection separates the
arms by itself: at $k = 0$ the context-conditioned completion is fully
adopted ($\lambda = 1$ on every fold and corpus) while the identical
context-free head is rejected outright ($\lambda = 0$, a~fallback to the
empirical table) on M3ED, EmoWOZ and 4 of 5 IEMOCAP folds. (iii)~The
gains are \emph{not} from more accurate completion prediction---the
learned completion's dev NLL is often worse than the empirical table's
(M3ED $k = 1$: $1.44$ vs.\ $1.39$)---but from row-specific,
context-conditioned weights that sharpen the target~marginalization.

At $k = 0$, the~total/context gains are $-0.065$/$-0.061$ nats on IEMOCAP,
$-0.095$/ $-0.074$ on M3ED, and~$-0.017$/$-0.015$ on EmoWOZ; the complete
three-term table with confidence intervals is in Appendix~\ref{app:ccprl}.

\subsection{Data-Amount Scan: When Consistency~Helps}
\label{sec:scan}

\begin{figure}[H]
\centering
\includegraphics[width=.999\textwidth]{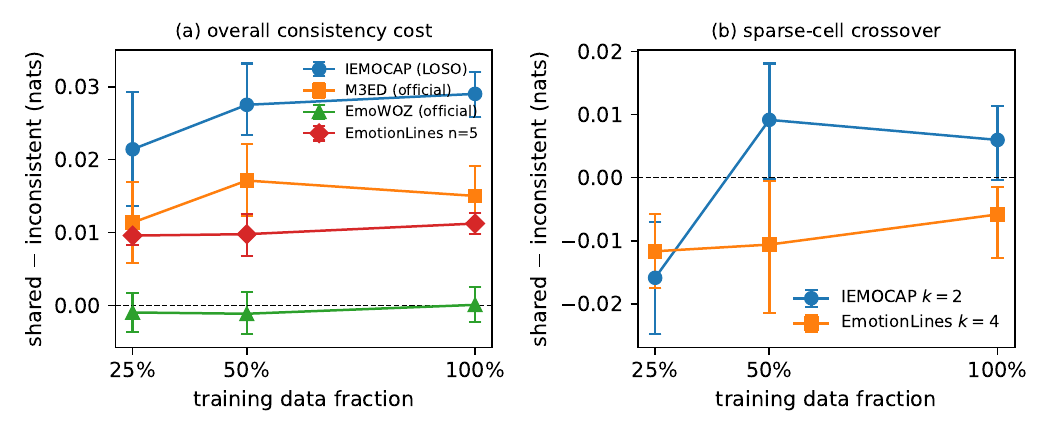}
\caption{Training data scan: shared consistent table minus per-level
inconsistent marginalizer (nats; positive = shared worse; error bars are
95\% CIs). (\textbf{a}) Overall cost shrinks as data shrinks. (\textbf{b}) Sparse cells flip:
IEMOCAP \texorpdfstring{$k = 2$}{k=2} and EmotionLines
\texorpdfstring{$n = 5$, $k = 4$}{n=5, k=4} favor the shared table as
per-level cells become sparser.}
\label{fig:scan}
\end{figure}

Per-level tables gain rows as $k$ grows: at $n = 5$, $k = 4$ indexes $210$
partial compositions ($420$ rows including relation) versus $42$ rows for
the shared summary table. This suggests a bias-variance reading of
the operator ablation in Appendix~\ref{app:operator_tables}:
per-level refits are low-bias but high-variance where cells are sparse,
while the shared table pools strength across levels and pays a bias cost
where per-level cells are dense. We test it by re-running the ablation on
deterministic row-level training subsamples at $25\%/50\%/100\%$ with the
same dev-selection protocol and seeds. Figure~\ref{fig:scan} and
Table~\ref{tab:scan} (Appendix~\ref{app:scan}) report the results: the
overall consistency cost shrinks with data scarcity (IEMOCAP
$0.029\to0.021$; EmotionLines $0.011\to0.010$; M3ED differences near zero), the~sparsest-cell advantage grows monotonically ($n = 5$, $k = 4$:
$-0.006\to-0.012$), mid-level cells flip (IEMOCAP $k = 2$:
$+0.006\to$ $-0.016$),
and~low-$k$ costs persist because those cells
stay dense even at $25\%$---exactly the crossover a bias-variance account
predicts. The~$100\%$ M3ED $k = 0/1$ values are $+0.041/+0.015$ in both
the main ablation table and this scan; they are read from the same
row-level JSON artifact rather than rounded~independently.

\subsubsection{CCPR-S}
The bias-variance picture suggests an operator that
adapts to cell density instead of committing to either endpoint. CCPR-S
does this with one scalar per level: it stays within $0.002$ nats of the
better endpoint everywhere, beats the shared table significantly on three
corpora, and~beats \emph{both} endpoints at the sparsest cells, while
keeping one shared table serving every $k$ and $n$
(Appendix~\ref{app:ccprs}).

\subsection{Rater-Count~Invariance}
\label{sec:ratercount}

The per-vote formulation lets one table serve any $n$. Zero-shot transfer
from 3-vote subsampled panels to full 5-rater panels wins $k = 0$ and
$k = 3,4,5$ with fold CIs excluding zero and is within $0.004$ nats at
$k = 1/2$; it holds on an external eval-gold set with no dialogue overlap
and is competitive with same-$n$ fitting despite never seeing $n = 5$ rows
(Appendix~\ref{app:n5budget}). In~a mixed $n \in \{3,5\}$ deployment a
single $n_0 = 3$ operator beats an oracle per-$n$ per-level lookup by
$-0.0187$ $[-0.0234,-0.0140]$, winning 9 of 10 $(n,k)$ cells; against the
stronger per-$(n,k)$ inconsistent marginalizer it costs $+0.0085$ nats,
halving to $+0.0044$ once that baseline is denied its oracle $n = 5$
training panels. Consistency is thus not predictively superior even
against an oracle refit; its value is zero-configuration service at a
small, fully quantified cost. At~$n = 5$ both models violate
$V(k + 1) \le V(k)$ at high $k$, a~finite-sample fitted-risk limitation in the
sparse 462-composition space that a per-level concentration probe does not
fix; we report it as a diagnostic case study of
Theorem~\ref{thm:monotone}.

\subsection{Identity Sensitivity: An Audit and a Stable-Subset~Check}
\label{sec:raterresolved}

The exchangeable operator deliberately represents a panel by counts rather
than by annotator identities. Before~fitting an identity-aware extension,
we audited whether the available files support that extension. The~old
IEMOCAP parser's three columns are not fixed evaluators: among the $7741$
strict exact-three utterances, five evaluator sets occur (the largest is
$e1/e2/e4$ with $6210$ rows). Thus the earlier claim that IEMOCAP ships
stable evaluator identities, and~the associated $0.85$-nat routing spread,
were not supported by the data representation and are~withdrawn.

As a sensitivity check, we reconstructed the actual evaluator IDs from the
categorical files and restricted both the current and previous utterance to
the common $e1/e2/e4$ subset ($3917$ rows, five sessions). An~identity-aware label table differs from a pooled table by only $+0.0074$
nats, with~a session bootstrap $95\%$ CI of
$[-0.0036,0.0190]$. This is evidence about a stable subset, not a claim
that all IEMOCAP rows share those three evaluators. M3ED and EmoWOZ expose
annotation slots without stable identities, and~EmotionLines supplies only
aggregated counts. We therefore do not claim that annotator routing is a
validated contribution of this revision. The~subset lattice identity
$V(S)-V(S\cup\{j\})=I(Y;R_j\mid X_S)$ remains a valid theoretical extension,
but its empirical deployment requires timestamped, identity-resolved
annotation logs that these corpora do not~provide.

\subsection{End-to-End with Neural~Context}
\label{sec:neural}

To show the anchor findings survive real context features, we train a
causal GRU Dirichlet-multinomial head with masked-rater augmentation (each
training row sees a random $k$ partial panel broadcast into the context)
and blend it with each anchor type in probability space (per-$k$ $\lambda$
on inner dev). On~IEMOCAP (text+visual), ccprJ+head beats lookup+head at
every level ($-0.012$/$-0.010$/$-0.005$/$-0.006$; CIs exclude zero, 5/5 at
every level). On~M3ED (CV folds, text+visual) the cold-start win replicates
($-0.063$, 5/5) while partial levels favor lookup+head by $\le 0.015$
nats (CIs exclude zero); on EmoWOZ (TF-IDF context) the mixture wins all
levels in CV ($-0.003$/$-0.002$/$-0.001$/$-0.001$) and official held-out
($-0.007$/$-0.004$/$-0.003$/$-0.007$). At~$n = 5$, ccprN+head wins
$k = 0$, $3, 4, 5$ ($-0.008$, $-0.012$, $-0.027$, $-0.031$; CIs exclude zero)
and is within $+0.007$ nats at $k = 1/2$. These mixtures are also the
context source for CCPR-L's completion head (Section~\ref{sec:learned}),
whose wins are the direct transfer of this context signal to the
completion, not a separate~effect.

\subsection{Annotation~Allocation}
\label{sec:budget}

Theorem~\ref{thm:cmi} supplies the population-level marginal-gain
interpretation, but~finite-sample fitted curves need not be monotone or
have non-increasing marginal gains. We therefore evaluate allocation
directly rather than assuming
that the fitted curve has the shape required by the greedy theorem. First,
a budget simulation on held-out
rows compares uniform allocation (a seeded, row-balanced
$\lfloor B \rfloor/\lceil B \rceil$ assignment) with concentrated allocation
($k = 3$ on a fraction $B/3$ and $k = 0$ elsewhere); each row's actual ccprJ
row NLL at its assigned level is used (no curve interpolation). Table~\ref{tab:budget}
(Appendix~\ref{app:n5budget}) reports the saving (concentrated $-$ uniform)
for $B \in \{0.5,\ldots,2.5\}$: uniform has positive expected saving in
all 15 cells, significant in~14.

Second, an~allocation policy needing no oracle. Rows split into disjoint
calibration and evaluation blocks at the session/series/dialogue level;
a greedy water-filling policy then awards each next rater to the largest
estimated marginal gain $\hat V_i(k)-\hat V_i(k + 1)$, which is the greedy
optimum when the population expected marginal gains are non-increasing
(Proposition~\ref{prop:greedy}). CCPR-A
estimates $\hat V$ by binning rows on the $k = 0$ forecast entropy---a
one-dimensional nearest-neighbor rule. CCPR-P instead \emph{amortizes}
the curve: a ridge regression on target-free row features (predicted mean
label vector and its entropy, top-1 composition mass, relation, forecast
entropy) predicts $\hat V_i(0..n)$. All features precede annotation, so
the policy is~deployable.

We also added simple target-free mean-label uncertainty controls: entropy,
least confidence, and~inverse margin of the predicted mean single-rater
label distribution. Each uses the same water-filling simulator and held-out
blocks, with~a fixed $1/(k + 1)$ schedule for later levels; they are
uncertainty proxies rather than full panel composition or universal active
learning algorithms. No heuristic dominates across corpora. At~$B = 1$ the
criteria are neutral on IEMOCAP by construction, least confidence is worse
on M3ED by $+0.0125$ nats, and~no criterion is significantly better than
uniform on EmoWOZ. At~$B = 2$, predictive entropy helps IEMOCAP by
$-0.0015$ nats, while least confidence and inverse margin hurt M3ED by
$+0.0402$ and $+0.0705$; the complete comparison is in
Appendix~\ref{app:heuristics}. These controls narrow the allocation claim
to the learned policy and do not support a universal acquisition~heuristic.

\subsubsection{What the Oracle Can and Cannot Bound}
Oracle-greedy forms
marginal gains from each row's \emph{realized} target, so it is
clairvoyant: part of its margin is target randomness no predictor can
recover, and~comparing against it conflates estimation error with
irreducible noise. We therefore add a \emph{retrospective feature class
headroom}: the same regression class fit in-sample on the evaluation rows
using their realized targets (\textsc{feature-bound}). This is a diagnostic
reference, not a deployment-time bound and not information available before
the evaluation outcomes are observed. Table~\ref{tab:active} reports both.
CCPR-P wins 10 of 15 cells against
uniform with CIs excluding zero (IEMOCAP 5/5, M3ED 5/5) versus 5 of 15
for CCPR-A (IEMOCAP 2/5, M3ED 3/5), with~$3$--$4\times$ larger effects, and~at $B = 1.0$ recovers
$83\%$/$93\%$ (IEMOCAP/M3ED) of the \emph{retrospective headroom} against
$23\%$/$26\%$ for binning ($35\%$/$41\%$ vs $10\%$/$11\%$ measured
against the clairvoyant oracle). On~EmoWOZ CCPR-P loses to uniform
(significantly at $B = 1.0$, $+0.021$), and~the bound diagnoses rather
than excuses it: \textsc{feature-bound} beats uniform by only $0.007$
nats there, so these features carry almost no row-level signal and the
regression fits noise. A~calibration-only cross-validation diagnostic is
the appropriate pre-deployment check; the retrospective reference is
reported solely to separate feature class headroom from clairvoyant target
randomness.

\subsection{Robustness and~Audit}
\label{sec:robust}

The iteration's $k = 3$ weight in $\{1,4,10\}$ leaves every conclusion unchanged
(IEMOCAP \linebreak  {$k = 0/1/2$ differences $-0.446/-0.092/-0.053$,
$-0.422$/$-0.075$/$-0.041$, $-0.412$/$-0.068$/}\linebreak  $-0.034$; all 5/5, CIs excluding
zero; the $k = 3$ cost stays in $0.021$--$0.031$). We retain fixed-order
revelation only as a sensitivity analysis, not as evidence about realistic
arrival. The~original files contain no annotation timestamps, and~the
row-wise evaluator order is not an identity-resolved arrival order. The~synthetic study therefore varies identity-dependent and
label-dependent masking explicitly. Historical implementation audits report passes for scripted
numeric (103/103), logic (26/26), CCPR-L (88/88), allocation (116/116), and~monotonicity (18/18 for the $n = 3$ suite) checks. These checks do not constitute independent replication; the $n = 5$ violation is
reported as a fitted-estimator~\mbox{diagnostic}.

\subsubsection{External-Style Baseline, and~What It Isolates} To the best of our
knowledge, no published system performs partial-revelation panel
forecasting. The~closest published modeling choice is the 
soft-label/distributional-ERC head: a softmax trained with cross-entropy against the mean label
distribution~\citep{wu2022disterc}. Instantiating it fairly needs care.
Scoring that head as a multinomial while the proper head keeps a
Dirichlet-multinomial confounds the training loss with the likelihood
family and inflates the gap by $0.39$ nats on EmoWOZ and $1.09$ on M3ED. Giving both heads the
same DM family, each with its own dev-selected concentration, collapses
the difference to $\le0.02$~nats (M3ED differences near zero; EmoWOZ favors the CE head
by $0.01$--$0.02$). In~these experiments, the~likelihood family accounts for
a much larger difference than the choice between the two training losses
(roughly $20$--$50$ times as large in the reported comparisons). This is a
modeling-choice observation, not a universal claim or a state-of-the-art claim.
Development selection favored a concentration scaling of $0.50$ for the
learned Dirichlet head on every evaluated corpus and level, indicating
that lower concentration improved NLL in this protocol.

\section{Negative Results and~Corrections}
\label{sec:negative}

Three claims made earlier in this work did not survive their own controls.
We state them together because each changes how a reader should interpret
a number that would otherwise look favorable, and~because the corrected
versions are more useful than the~\mbox{originals.}

\subsection{Exact-Marginal Training of the Completion Head Fails} The
completion head is trained with cross-entropy against the observed
completion, which is a surrogate: it is minimized by accurate completion
prediction, not by accurate target forecasting, and~the two demonstrably
come apart (the learned completion's development NLL is often worse than
the empirical table's while its target NLL is better). Because~the
completion latent is enumerable, the~exact marginal objective
$-\log\sum_m \mathrm{DM}(c_t\mid\kappa T_{g(\tilde c+m)})\,W_\theta(m\mid x)$
is available in closed form and differentiable, so we trained the head on
it directly. It is \emph{worse}: $+0.064/+0.043/+0.015$ nats at
$k = 0/1/2$ on IEMOCAP relative to the surrogate-trained head, and~development selection discarded the exact-marginal head outright
($\lambda = 0$) on four of five folds. Freed of the surrogate's semantic
anchor, the~completion weights become an unconstrained nuisance parameter
that overfits. The~general guidance that exact marginalization beats
approximate surrogates does not transfer here; the surrogate is doing
necessary regularization~work.

\subsection{The Dirichlet-Versus-Cross-Entropy Comparison Was a
Likelihood-Family Artifact} A soft-label head trained with cross-entropy
against the mean label distribution appeared substantially worse than a
proper Dirichlet-multinomial head. That comparison scored the
cross-entropy head as a multinomial while the proper head kept a
Dirichlet-multinomial, confounding the training loss with the likelihood
family: on M3ED the family alone is worth $1.09$ nats and on EmoWOZ
$0.39$. Giving both heads the same Dirichlet-multinomial family, each with
its own development-selected concentration, collapses the difference to
$\le 0.02$ nats. The~defensible conclusion is narrower than the original
one: in this object and protocol, the~likelihood family contributes much
more to the observed difference than the training loss choice. Development selection also favored halving the learned Dirichlet
concentration on every evaluated corpus and level. This improves NLL in
this protocol; it does not quantify miscalibration as a factor of two.

\subsection{The Usual Allocation Oracle Bounds Nothing Deployable} Policies
that allocate annotation are conventionally compared against an oracle
that forms marginal gains from each item's \emph{realized} label. That
oracle is clairvoyant: part of its margin is target randomness that no
predictor can recover, so ``fraction of the oracle recovered'' conflates
estimation error with irreducible noise. We therefore report both the
clairvoyant oracle and a retrospective feature class headroom---the same
model class fit in-sample on the evaluation block. The~latter uses observed
evaluation targets and is not a deployable upper bound. The~same policy
recovers $35$--$41\%$ of the clairvoyant oracle and $83$--$93\%$ of this
retrospective headroom on IEMOCAP/M3ED; EmoWOZ supplies no positive
headroom~result.

\subsection{Two Further Nulls, Reported for Completeness} Coarsening
consistency loses to a per-level marginalizer even when that baseline is
denied its oracle panel-size training data (cost $+0.0085$ nats, halving
to $+0.0044$). The~identity-aware extension is also not supported as a
general empirical contribution: the original IEMOCAP columns are not
stable IDs, and~the stable-subset check is compatible with no identity
effect.

\section{Discussion: What This Means for Annotation~Projects}
\label{sec:discussion}

Three practical consequences follow, in~decreasing order of how confident
we are in~them.

\subsection{Spread the Budget; Measure the Gain Rather than Assume Its Shape}
The held-out budget simulation finds positive saving for uniform allocation
over concentrated allocation in all 15 budget cells and significant saving
in 14. This operational result is more defensible than a universal
diminishing-return claim: fitted curves can violate the population shape at
sparse levels, as~the $n = 5$ diagnostic~shows.

\subsection{Practitioner Takeaway} A second same-item label has measurable but
diminishing predictive value in all three direct-validation corpora: the
first/second additional votes reduce NLL by $0.583/0.221$ nats on IEMOCAP,
$0.578/0.184$ on M3ED, and~$0.180/0.069$ on EmoWOZ. When the objective is
broad panel coverage, the~held-out budget simulation instead favors
spreading a fixed budget across items rather than concentrating it on a
smaller set. These are protocol- and corpus-specific operational findings,
not a universal rule for every annotation~project.

\subsection{Predicting Per-Item Value Can Help, but~the Reference Must Be
Honest} An amortized policy that regresses each item's curve on
target-free features and then water-fills greedily recovers much of the
retrospective feature class headroom where the features carry item-level
signal, and~loses where they do not. That headroom is fit using evaluation
targets and is therefore retrospective, not a pre-deployment ceiling. A~practitioner should use a calibration-only cross-validation diagnostic to
decide whether adaptive allocation is worth building; the EmoWOZ result
illustrates why that check~matters.

\subsection{Identity-Aware Routing Remains an Open Problem} The theoretical
subset lattice extension is straightforward, but~the corpora do not provide
the timestamped, identity-resolved arrival logs needed to evaluate it
cleanly. An~audit of IEMOCAP shows that the row-wise three-column order is
not a fixed evaluator identity; the common stable subset yields only
$+0.0074$ nats relative to a pooled table with a CI crossing zero. We
therefore remove the earlier routing success claim and leave identity-aware
allocation for datasets with an appropriate annotation~log.

\subsection{Modeling Guidance} For practitioners fitting panel
forecasters: choose the likelihood family carefully (the
Dirichlet-multinomial versus multinomial contrast is much larger than the
training loss contrast in these experiments); calibrate the concentration
post-hoc; and prefer a shared consistent table when annotation levels or
panel sizes are numerous relative to data, a~per-level fit when cells are
dense.

\section{Limitations}
\label{sec:limitations}

\begin{enumerate}
\item \textbf{Gains are corpus-dependent.}
Cold-start wins replicate on all
corpora; mid-revelation gains are significant on IEMOCAP and EmoWOZ,
statistically indistinguishable from zero on M3ED, and~slightly negative
for some neural mixtures; we report every cell.
\item \textbf{Consistency does not guarantee predictive improvement.} The per-level
inconsistent marginalizer matches or beats the consistent operator at
every level (significantly at low $k$ on IEMOCAP and M3ED; near-zero differences on
EmoWOZ), and~in mixed-$\{3,5\}$ deployment an oracle per-$(n,k)$ refit
beats the single shared table by $+0.0085$ nats overall (CI
$[0.005,0.012]$), although~the shared table wins the sparsest cells.
Consistency is a structured regularizer---its cost concentrates in dense
cells and shrinks, sometimes flipping, under~data scarcity. Its value is
zero-configuration single-table deployment and auditability at a
quantified cost.
\item \textbf{CCAR is an assumption.} CCAR is a controlled masking
intervention, not an empirically verified arrival process. The~raw corpora
do not contain the timestamps needed to reconstruct arrival order; our
identity- and label-dependent masking results are synthetic sensitivity
analyses, not measurements of the original annotation workflow.
\item \textbf{Small and domain-specific evidence.} The corpora are
conversational emotion datasets, and~IEMOCAP has only five independent
sessions. The~main real-data evaluations use panels of $n = 3$ and $n = 5$;
the behavior of the operators for substantially larger real-world panels
remains to be established. Same-item validation is therefore evidence for
this setting, not a universal annotation-budgeting law; session/block effects
are reported alongside effect sizes.
\item \textbf{No leaderboard comparison.} Baselines are same-protocol,
capacity-matched internal methods; we make no state-of-the-art claim. Datasets:
IEMOCAP (licensed copy), M3ED (ACL 2022 release), EmoWOZ (CC BY-NC 4.0,
research use only).
\end{enumerate}

\section{Conclusions}
\label{sec:conclusion}

We formalized partial-revelation panel forecasting and separated two
estimands that had previously been conflated. For~the cross-turn task,
the population decrement is a conditional mutual information; for a
fitted forecaster, the~decrement also contains adjacent excess predictive
risk. A~held-out rater experiment measures same-item predictive value
directly: the first/second additional votes reduce NLL by
$0.583/0.221$ nats on IEMOCAP, $0.578/0.184$ on M3ED, and~$0.180/0.069$ on EmoWOZ. These results support additional-label
informativeness in this setting, but~do not make cross-turn gain a
universal synonym for annotation~value.

On the modeling side we defined coarsening-consistent panel operators,
gave a closed-form Dirichlet-multinomial realization, and~characterized
consistency honestly: it is a coherence and deployment property, not an
empirical source of predictive accuracy. The~allocation experiments show
that uniform allocation has a positive estimated saving over the tested
concentrated policy in all 15 held-out cells, significant in 14 under the
prespecified block-bootstrap comparisons; this is evidence for the tested
protocol, not a universal allocation law. Target-free adaptive allocation
helps only when row-level signal is predictable. Its $83$--$93\%$ figure is relative to a
retrospective feature class headroom, not a deployable or population upper
bound; the clairvoyant-oracle comparison is $35$--$41\%$ on the reported
IEMOCAP/M3ED setting. Identity-aware routing is left open because the
available data do not support a clean arrival-and-identity~evaluation.

Section~\ref{sec:negative} reports negative results and the controls
that constrain their interpretation. The ancillary artifact documents
implementation scripts, stored aggregate outputs, and the checks performed
on the reported summaries. Fixed seeds and these checks support inspection,
but do not by themselves certify bit-identical reproduction across
environments or the absence of every possible data-processing error.

\vspace{6pt}

\authorcontributions{Conceptualization, B.W. and T.-P.T.; methodology,
B.W.; software, B.W.; validation, B.W. and D.-Q.Z.; formal analysis, B.W.
and D.-Q.Z.; investigation, B.W.; writing---original draft preparation,
B.W.; writing---review and editing, B.W., D.-Q.Z. and T.-P.T.;
supervision, T.-P.T. All authors have read and approved the manuscript.}

\funding{This research received no external~funding.}

\institutionalreview{Not applicable. This study analyses existing,
previously collected and publicly documented annotation corpora and did
not involve any new data collection from humans or animals.}

\informedconsent{Not applicable.}

\dataavailability{This study uses IEMOCAP (USC academic licence), M3ED
(ACL 2022 release), EmoWOZ DialMAGE (CC BY-NC 4.0), and EmotionLines.
The original corpora and per-item annotations are not redistributed here;
readers must obtain them under their respective terms. The ancillary
materials accompanying this preprint
contain implementation scripts, additional validation scripts, stored
aggregate result summaries, and an artifact manifest. The public repository
\url{https://github.com/xiukkie/cpr-panel-forecasting} contains an earlier
implementation and is not a complete mirror of the current revision.
Rebuilding training runs requires the original datasets and, for the
multimodal experiments, the source features described by the scripts.
The included aggregate summaries permit inspection of reported results;
they do not substitute for a full independent training reproduction.}

\acknowledgments{The authors used AI-based assistive tools for language editing and,
in preparing this preprint, for mathematical consistency review, reference
checks, \LaTeX{} preparation, and software-assisted checks of stored result
summaries. The authors remain responsible for the research, interpretations,
and all contents of the manuscript.}
\conflictsofinterest{The authors declare no conflicts of~interest.}

\appendix
\section[\appendixname~\thesection]{Proofs}\label{app:proofs}

\begin{Lemma}[DM closure]
\label{lem:dm}
Let $\theta\sim\mathrm{Dirichlet}(\beta)$ with $\beta_j>0$, and let the
votes be conditionally independent categorical draws given $\theta$. Under
an ignorable CCAR revelation subset, observing partial counts
$\tilde c^{(k)}$ leaves the $n-k$ unobserved counts $m$ distributed as
$\mathrm{DM}_{n-k}(m\mid\beta+\tilde c^{(k)})$, by conjugacy.
Exchangeability makes this result independent of which subset is revealed.
\end{Lemma}

\begin{Lemma}[Mixture closure]
\label{lem:mixture}
If $\{Q_k\}$ and $\{Q'_k\}$ are consistent families sharing the same
completion posterior, then for every fixed $\lambda \in [0,1]$,
held constant across revelation levels, the mixture
$R_k = (1-\lambda)Q_k + \lambda Q'_k$ satisfies
\eqref{eq:consistency}, by~linearity of expectation.
\end{Lemma}

\begin{proof}[Proof of Theorem~\ref{thm:monotone}]
Because the revelation process is nested,
$\sigma(X_k)\subseteq\sigma(X_{k+1})$ and the tower property gives
$P_k=\mathbb{E}[P_{k+1}\mid X_k]$. For~each
realization $(x_k, r)$, strict propriety of $S$ gives
$\mathbb{E}[S(P(Y\mid x_k,r), Y) \mid x_k, r] \le
\mathbb{E}[S(P(Y\mid x_k), Y) \mid x_k, r]$, because~$P(Y\mid x_k)$ is the
Bayes act for the coarser information while $P(Y\mid x_k,r)$ is the finer
forecast. Taking expectations over $R \mid X_k$ and then $X_k$ yields the
claim. The~logarithmic score (count NLL) is strictly proper for
full-support distributions over count vectors.
\end{proof}

\begin{proof}[Proof of Theorem~\ref{thm:cmi}]
Write
\begin{align}
V(k) - V(k+1)
&= \mathbb{E}\bigl[\log P(Y\mid X_{k+1}) - \log P(Y\mid X_k)\bigr]
\nonumber\\
&= \mathbb{E}_{X_k}\Bigl[\mathbb{E}_{R \mid X_k}\bigl[
\mathrm{KL}\bigl(P(Y\mid X_k, R) \,\|\, P(Y\mid X_k)\bigr)\bigr]\Bigr]
\nonumber\\
&= I(Y; R_{k+1} \mid X_k).
\end{align}
Summing over $k$ telescopes: $V(0)-V(n) = I(Y; X_n\mid X_0)$.
The unconditional form requires $I(Y;X_0)=0$.
\end{proof}

\begin{Proposition}[Consistency gap of per-level lookup]
\label{prop:gap}
The per-level lookup family is consistent only if each level-$k$ table
equals the completion-marginalized full-panel table; otherwise the KL gap
in \eqref{eq:gap} is positive. Empirically the gap is $0.02$--$0.05$ (IEMOCAP
$k = 1/2$) versus $\approx 10^{-17}$ for CCPR-J and $0.003$--$0.008$ for
CCPR-D.
\end{Proposition}

\begin{Proposition}[Shrinkage gap bound]
\label{prop:gapmix}
Let $Q^{\mathrm{shared}}$ and $Q^{\mathrm{per-level}}$ be two panel-operator
families sharing a common completion kernel, with per-level consistency
gaps $\mathrm{Gap}_{\mathrm{s}}(k)$ and $\mathrm{Gap}_{\mathrm{p}}(k)$ measured
using that kernel. For every fixed $\lambda\in[0,1]$ held constant across
revelation levels, the mixture
$Q^\lambda = (1-\lambda)Q^{\mathrm{shared}} + \lambda Q^{\mathrm{per-level}}$
satisfies $\mathrm{Gap}_{\lambda}(k) \le (1-\lambda)\mathrm{Gap}_{\mathrm{s}}(k)
+ \lambda \mathrm{Gap}_{\mathrm{p}}(k)$, by joint convexity of KL, because
the mixed full-panel reference is the same convex combination of the two
endpoint references. The claim does not cover level-specific mixing weights
or different completion kernels.
\end{Proposition}

\section[\appendixname~\thesection]{Full NLL~Tables}
\label{app:tables}

\begin{table}[H]
\centering
\caption{IEMOCAP count NLL by revelation level (session LOSO).}
\label{tab:ie}
\newcolumntype{C}{>{\centering\arraybackslash}X}
\newcolumntype{B}{>{\raggedright\arraybackslash}X}
\newcolumntype{A}{>{\raggedleft\arraybackslash}X}
\begin{tabularx}{\textwidth}{CCCCC}

\toprule

\boldmath{$k$} & \textbf{ccprJ} & \textbf{ccprD} & \textbf{Lookup} & \textbf{Flat} \\
\midrule
0 & 3.2482 & 3.2482 & 3.6703 & 3.6836 \\
1 & 2.6903 & 2.6903 & 2.7654 & 3.6341 \\
2 & 2.5446 & 2.5446 & 2.5854 & 3.6341 \\
3 & 2.5406 & 2.5195 & 2.5195 & 3.6341 \\
\bottomrule

\end{tabularx}

\end{table}
\unskip

\begin{table}[H]
\centering
\caption{M3ED official count~NLL.}
\label{tab:m3}
\newcolumntype{C}{>{\centering\arraybackslash}X}
\newcolumntype{B}{>{\raggedright\arraybackslash}X}
\newcolumntype{A}{>{\raggedleft\arraybackslash}X}
\begin{tabularx}{\textwidth}{CCCc}

\toprule
\boldmath{$k$} & \textbf{ccprJ} & \textbf{Lookup} &\textbf{ccprJ} \boldmath{$-$} \textbf{Lookup} \\
\midrule
0 & 2.5266 & 2.6925 & $-0.166$ $[-0.206,-0.125]$ \\
1 & 1.9721 & 1.9665 & $+0.006$ $[-0.005,0.017]$ \\
2 & 1.7192 & 1.7121 & $+0.007$ $[-0.000,0.014]$ \\
3 & 1.6002 & 1.5997 & $+0.001$ $[-0.011,0.012]$ \\
\bottomrule

\end{tabularx}
\end{table}
\unskip

\begin{table}[H]
\centering
\caption{EmoWOZ official count~NLL.}
\label{tab:em}
\newcolumntype{C}{>{\centering\arraybackslash}X}
\newcolumntype{B}{>{\raggedright\arraybackslash}X}
\newcolumntype{A}{>{\raggedleft\arraybackslash}X}
\begin{tabularx}{\textwidth}{CCCc}

\toprule
\boldmath{$k$} & \textbf{ccprJ} & \textbf{Lookup} &\textbf{ccprJ} \boldmath{$-$} \textbf{Lookup} \\
\midrule
0 & 2.0429 & 2.0565 & $-0.014$ $[-0.023,-0.003]$ \\
1 & 1.9573 & 1.9636 & $-0.006$ $[-0.011,-0.001]$ \\
2 & 1.9318 & 1.9373 & $-0.006$ $[-0.010,-0.001]$ \\
3 & 1.9158 & 1.9183 & $-0.002$ $[-0.010,0.006]$ \\
\bottomrule

\end{tabularx}
\end{table}

\section[\appendixname~\thesection]{Operator Ablation~Details}
\label{app:operator_tables}

The following table contains the full consistency-ablation values moved from
the main text to keep the core narrative focused on the two estimands and
the allocation~result.

\begin{table}[H]
\centering
\small
\caption{Consistency ablation: ccprJ-eq \texorpdfstring{$-$}{-} per-level inconsistent
marginalizer (nats). Positive = shared table worse. Bold: CI excludes zero.}
\label{tab:abl}
\newcolumntype{C}{>{\centering\arraybackslash}X}
\newcolumntype{B}{>{\raggedright\arraybackslash}X}
\newcolumntype{A}{>{\raggedleft\arraybackslash}X}
\begin{tabularx}{\textwidth}{lCCCC}

\toprule

\textbf{Corpus} & \boldmath{$k = 0$} & \boldmath{$k = 1$} & \boldmath{$k = 2$} & \boldmath{$k = 3$} \\
\midrule
IEMOCAP & \boldmath{$+0.048$} & \boldmath{$+0.034$} & $+0.006$ & \boldmath{$+0.028$} \\
 & $[0.044,0.052]$ & $[0.029,0.038]$ & $[-0.000,0.011]$ & $[0.017,0.042]$ \\
M3ED official & \boldmath{$+0.041$} & \boldmath{$+0.015$} & $+0.003$ & $+0.000$ \\
 & $[0.029,0.054]$ & $[0.009,0.022]$ & $[-0.002,0.008]$ & $[-0.008,0.008]$ \\
EmoWOZ official & $+0.004$ & $-0.001$ & $-0.002$ & $-0.001$ \\
 & $[-0.004,0.011]$ & $[-0.004,0.002]$ & $[-0.005,0.001]$ & $[-0.005,0.004]$ \\
\bottomrule

\end{tabularx}

\end{table}
\unskip

\section[\appendixname~\thesection]{Data-Amount Scan~Details}
\label{app:scan}

The scan subsamples training rows deterministically at each fraction
(seed per fold and level), re-runs the same dev selection and the same posterior-weighted
moment fits, and~evaluates on the identical held-out rows as the full-data run.
For $n = 3$ corpora the shared model is ccprJ-eq (equal $k$-weights); for
EmotionLines it is the shared CCPR-N summary table. Delta is shared minus
the per-level inconsistent marginalizer (positive = shared worse); bold
cells have 95\% CIs excluding zero (session/fold bootstrap for IEMOCAP and
EmotionLines; dialogue bootstrap for M3ED and EmoWOZ).

\begin{table}[H]
\centering
\small
\caption{Training data scan: shared \texorpdfstring{$-$}{-} inconsistent (nats), by~corpus,
training fraction, and~revelation level; overall is the equal-weight mean
over $k$. Bold indicates cells whose 95\% bootstrap CIs exclude zero
(session/fold bootstrap for IEMOCAP and EmotionLines; dialogue bootstrap
for M3ED and EmoWOZ).}
\label{tab:scan}
\newcolumntype{C}{>{\centering\arraybackslash}X}
\newcolumntype{B}{>{\raggedright\arraybackslash}X}
\newcolumntype{A}{>{\raggedleft\arraybackslash}X}
\begin{tabularx}{\textwidth}{lCCCCCCCC}

\toprule

\textbf{Corpus} & \textbf{Frac} & \boldmath{$k = 0$} & \boldmath{$k = 1$} & \boldmath{$k = 2$} & \boldmath{$k = 3$} & \boldmath{$k = 4$} & \boldmath{$k = 5$} & \textbf{Overall} \\
\midrule
IEMOCAP & 25\% & \boldmath{$+0.042$} & \boldmath{$+0.025$} & \boldmath{$-0.016$} & \boldmath{$+0.035$} & --- & --- & \boldmath{$+0.021$} \\
 & 50\% & \boldmath{$+0.046$} & \boldmath{$+0.025$} & $+0.009$ & \boldmath{$+0.030$} & --- & --- & \boldmath{$+0.028$} \\
 & 100\% & \boldmath{$+0.048$} & \boldmath{$+0.034$} & $+0.006$ & \boldmath{$+0.028$} & --- & --- & \boldmath{$+0.029$} \\
M3ED official & 25\% & \boldmath{$+0.034$} & \boldmath{$+0.007$} & $-0.003$ & $+0.007$ & --- & --- & \boldmath{$+0.011$} \\
 & 50\% & \boldmath{$+0.040$} & \boldmath{$+0.009$} & \boldmath{$+0.006$} & \boldmath{$+0.013$} & --- & --- & \boldmath{$+0.017$} \\
 & 100\% & \boldmath{$+0.041$} & \boldmath{$+0.015$} & $+0.003$ & $+0.000$ & --- & --- & \boldmath{$+0.015$} \\
EmoWOZ official & 25\% & $+0.005$ & $-0.003$ & $-0.002$ & $-0.003$ & --- & --- & $-0.001$ \\
 & 50\% & $+0.002$ & $-0.002$ & $-0.003$ & $-0.003$ & --- & --- & $-0.001$ \\
 & 100\% & $+0.004$ & $-0.001$ & $-0.002$ & $-0.001$ & --- & --- & $+0.000$ \\
EmotionLines $n = 5$ & 25\% & \boldmath{$+0.028$} & \boldmath{$+0.023$} & \boldmath{$+0.014$} & $+0.005$ & \boldmath{$-0.012$} & $0.000$ & \boldmath{$+0.010$} \\
 & 50\% & \boldmath{$+0.029$} & \boldmath{$+0.022$} & \boldmath{$+0.015$} & $+0.003$ & \boldmath{$-0.011$} & $0.000$ & \boldmath{$+0.010$} \\
 & 100\% & \boldmath{$+0.032$} & \boldmath{$+0.023$} & \boldmath{$+0.014$} & \boldmath{$+0.004$} & \boldmath{$-0.006$} & $0.000$ & \boldmath{$+0.011$} \\
\bottomrule

\end{tabularx}

\end{table}

\section[\appendixname~\thesection]{Rater-Count Transfer and Budget~Tables}
\label{app:n5budget}

\begin{table}[H]
\centering
\small
\caption{Mixed-\texorpdfstring{$\{3,5\}$}{3,5} deployment: shared table minus
per-\texorpdfstring{$(n,k)$}{(n,k)}
inconsistent marginalizer, by~cell (nats; positive = shared worse).}
\label{tab:mixed}
\newcolumntype{C}{>{\centering\arraybackslash}X}
\newcolumntype{B}{>{\raggedright\arraybackslash}X}
\newcolumntype{A}{>{\raggedleft\arraybackslash}X}
\begin{tabularx}{\textwidth}{BCCCCCC}

\toprule

\textbf{Cell} & \boldmath{$k = 0$} & \boldmath{$k = 1$} & \boldmath{$k = 2$} & \boldmath{$k = 3$} & \boldmath{$k = 4$} & \boldmath{$k = 5$} \\
\midrule
$n = 3$ & $+0.009$ & $+0.009$ & $-0.003$ & $0.000$ & --- & --- \\
$n = 5$ & $+0.025$ & $+0.025$ & $-0.000$ & $+0.018$ & $-0.002$ & $+0.012$ \\
\bottomrule

\end{tabularx}

\end{table}
\unskip

\begin{table}[H]
\centering
\small
\caption{CCPR-N minus the in-corpus per-level lookup at the test rater
count, on~EmotionLines (nats). Bold: fold-bootstrap CI excludes zero.}
\label{tab:n5}
\newcolumntype{C}{>{\centering\arraybackslash}X}
\newcolumntype{B}{>{\raggedright\arraybackslash}X}
\newcolumntype{A}{>{\raggedleft\arraybackslash}X}
\begin{tabularx}{\textwidth}{lCCCCCC}

\toprule
\textbf{Condition} & \boldmath{$k = 0$} & \boldmath{$k = 1$} & \boldmath{$k = 2$} & \boldmath{$k = 3$} & \boldmath{$k = 4$} & \boldmath{$k = 5$} \\
\midrule
$n = 3\to n = 5$ (zero-shot) & \boldmath{$-0.011$} & $+0.001$ & $-0.003$ & \boldmath{$-0.016$} & \boldmath{$-0.039$} & \boldmath{$-0.041$} \\
$n = 5\to n = 5$ & \boldmath{$-0.012$} & $+0.001$ & \boldmath{$-0.010$} & \boldmath{$-0.036$} & \boldmath{$-0.084$} & \boldmath{$-0.087$} \\
$n = 5\to n = 3$ & \boldmath{$-0.008$} & $+0.000$ & \boldmath{$-0.004$} & $+0.006$ & --- & --- \\
$n = 3\to n = 3$ & \boldmath{$-0.007$} & $+0.002$ & $-0.004$ & $+0.017$ & --- & --- \\
\bottomrule

\end{tabularx}
\end{table}
\unskip

\begin{table}[H]
\centering
\caption{Annotation budget: saving of uniform over concentrated allocation
(nats/step; concentrated \texorpdfstring{$-$}{-} uniform). Bold: CI excludes zero.}
\label{tab:budget}
\newcolumntype{C}{>{\centering\arraybackslash}X}
\newcolumntype{B}{>{\raggedright\arraybackslash}X}
\newcolumntype{A}{>{\raggedleft\arraybackslash}X}
\begin{tabularx}{\textwidth}{lCCCCC}

\toprule
\textbf{Corpus} & \boldmath{$B = 0.5$} & \boldmath{$B = 1.0$} & \boldmath{$B = 1.5$} & \boldmath{$B = 2.0$} & \boldmath{$B = 2.5$} \\
\midrule
IEMOCAP & \boldmath{$+0.168$} & \boldmath{$+0.341$} & \boldmath{$+0.297$} & \boldmath{$+0.239$} & \boldmath{$+0.109$} \\
M3ED official & \boldmath{$+0.133$} & \boldmath{$+0.248$} & \boldmath{$+0.208$} & \boldmath{$+0.189$} & \boldmath{$+0.088$} \\
EmoWOZ official & $+0.015$ & \boldmath{$+0.042$} & \boldmath{$+0.041$} & \boldmath{$+0.030$} & \boldmath{$+0.031$} \\
\bottomrule

\end{tabularx}
\end{table}
\unskip

\begin{table}[H]
\centering
\small
\setlength{\tabcolsep}{3pt}
\caption{CCPR-P: target-free learned water-filling policy versus uniform,
per-row mean count NLL. ``Retrospective'' is the in-sample feature class
reference and is not deployable; oracle greedy uses the evaluation rows'
realized curves. The~final column gives CCPR-P minus uniform with
block-bootstrap CIs (positive = CCPR-P worse). Final-column entries are
bold when their 95\% block-bootstrap CIs exclude zero.}
\label{tab:active}
\newcolumntype{C}{>{\centering\arraybackslash}X}
\newcolumntype{B}{>{\raggedright\arraybackslash}X}
\newcolumntype{A}{>{\raggedleft\arraybackslash}X}
\begin{tabularx}{\textwidth}{lcCCCCc}

\toprule
\textbf{Corpus} & \boldmath{$B$} & \textbf{CCPR-P} & \textbf{Uniform} & \textbf{Retrospective} & \textbf{Oracle} & \textbf{P} \boldmath{$-$} \textbf{Uniform} \\
\midrule
IEMOCAP & 0.5 & 2.4132 & 2.4812 & 2.3865 & 2.2462 & \boldmath{$-0.068$ $[-0.093,-0.049]$} \\
 & 1.0 & 2.3511 & 2.4449 & 2.3317 & 2.1774 & \boldmath{$-0.094$ $[-0.135,-0.057]$} \\
 & 1.5 & 2.3381 & 2.4166 & 2.3207 & 2.1612 & \boldmath{$-0.079$ $[-0.111,-0.057]$} \\
 & 2.0 & 2.3500 & 2.3850 & 2.3202 & 2.1786 & \boldmath{$-0.035$ $[-0.053,-0.009]$} \\
 & 2.5 & 2.3487 & 2.3630 & 2.3337 & 2.2303 & \boldmath{$-0.014$ $[-0.023,-0.005]$} \\
M3ED & 0.5 & 2.0357 & 2.1280 & 2.0066 & 1.7877 & \boldmath{$-0.092$ $[-0.150,-0.024]$} \\
 & 1.0 & 1.8025 & 1.9216 & 1.7936 & 1.6324 & \boldmath{$-0.119$ $[-0.151,-0.077]$} \\
 & 1.5 & 1.6967 & 1.8032 & 1.6443 & 1.5577 & \boldmath{$-0.107$ $[-0.127,-0.081]$} \\
 & 2.0 & 1.6000 & 1.6840 & 1.5616 & 1.5043 & \boldmath{$-0.084$ $[-0.126,-0.039]$} \\
 & 2.5 & 1.5948 & 1.6415 & 1.5612 & 1.4838 & \boldmath{$-0.047$ $[-0.071,-0.023]$} \\
EmoWOZ & 0.5 & 1.9019 & 1.8947 & 1.8821 & 1.7811 & $+0.007$ $[-0.012,+0.026]$ \\
 & 1.0 & 1.8812 & 1.8606 & 1.8538 & 1.7371 & \boldmath{$+0.021$ $[0.006,0.037]$} \\
 & 1.5 & 1.8617 & 1.8513 & 1.8379 & 1.7226 & $+0.010$ $[-0.008,+0.029]$ \\
 & 2.0 & 1.8461 & 1.8382 & 1.8256 & 1.7264 & $+0.008$ $[-0.012,0.027]$ \\
 & 2.5 & 1.8291 & 1.8191 & 1.8195 & 1.7478 & $+0.010$ $[-0.003,+0.024]$ \\
\bottomrule

\end{tabularx}
\end{table}

\section[\appendixname~\thesection]{Standard Allocation~Heuristics}
\label{app:heuristics}

The following table reports simple target-free mean-label uncertainty
controls under the same calibration/evaluation split and water-filling
budget simulator as CCPR-P. Entries are policy minus uniform in mean count
NLL; negative is better. The~scores are computed from the predicted mean
single-rater label distribution, not the full panel composition distribution,
and use a fixed $1/(k + 1)$ schedule at later levels. Predictive entropy is
exactly uniform at $B = 1$ because one additional vote is assigned to every
row.

\begin{table}[H]
\centering
\caption{Mean-label uncertainty allocation controls. Entries are
policy minus uniform with 95\% block-bootstrap CIs (nats; negative is
better).}
\label{tab:heuristics}
\newcolumntype{C}{>{\centering\arraybackslash}X}
\newcolumntype{B}{>{\raggedright\arraybackslash}X}
\newcolumntype{A}{>{\raggedleft\arraybackslash}X}
\begin{tabularx}{\textwidth}{Blcc}

\toprule

\textbf{Corpus} & \textbf{Policy} & \boldmath{$B = 1.0$} & \boldmath{$B = 2.0$} \\
\midrule
IEMOCAP & predictive entropy & $0.000$ $[0.000,0.000]$ & $-0.0015$ $[-.0021,-.0007]$ \\
 & least confidence & $-0.0037$ $[-.0091,.0013]$ & $+0.0011$ $[-.0062,.0067]$ \\
 & inverse margin & $-0.0019$ $[-.0055,.0029]$ & $+0.0167$ $[.0036,.0312]$ \\
M3ED & predictive entropy & $0.000$ $[0.000,0.000]$ & $+0.0009$ $[-.0019,.0040]$ \\
 & least confidence & $+0.0125$ $[.0074,.0210]$ & $+0.0402$ $[.0286,.0579]$ \\
 & inverse margin & $+0.0702$ $[.0288,.1212]$ & $+0.0705$ $[.0377,.1001]$ \\
EmoWOZ & predictive entropy & $0.000$ $[0.000,0.000]$ & $-0.0007$ $[-.0025,.0009]$ \\
 & least confidence & $-0.0018$ $[-.0157,.0097]$ & $+0.0004$ $[-.0099,.0098]$ \\
 & inverse margin & $+0.0133$ $[-.0147,.0373]$ & $+0.0033$ $[-.0148,.0201]$ \\
\bottomrule

\end{tabularx}

\end{table}

\section[\appendixname~\thesection]{CCPR-S Per-Level~Table}
\label{app:ccprs}

\begin{table}[H]
\centering
\small
\caption{CCPR-S per-level NLL and deltas (nats; positive = ccprS worse;
bold = 95\% CI excludes zero). Shared = ccprJ-eq for
\texorpdfstring{$n = 3$}{n=3}, ccprN for \texorpdfstring{$n = 5$}{n=5}.}
\label{tab:ccprs_full}
\newcolumntype{C}{>{\centering\arraybackslash}X}
\newcolumntype{B}{>{\raggedright\arraybackslash}X}
\newcolumntype{A}{>{\raggedleft\arraybackslash}X}
\begin{tabularx}{\textwidth}{lcCCcCCc}

\toprule

\textbf{Corpus} & \textbf{Level} & \textbf{ccprS} & \textbf{Shared} & \textbf{Per-Level} & \textbf{Lookup} & \textbf{S}\boldmath{$-$}\textbf{Inc} & \textbf{S}\boldmath{$-$}\textbf{Shared} \\
\midrule
IEMOCAP & $k = 0$ & 3.1765 & 3.2244 & 3.1765 & 3.6703 & $+0.000$ & \boldmath{$-0.048$} \\
 & $k = 1$ & 2.6395 & 2.6729 & 2.6390 & 2.7654 & \boldmath{$+0.001$} & \boldmath{$-0.034$} \\
 & $k = 2$ & 2.5256 & 2.5325 & 2.5265 & 2.5854 & $-0.001$ & \boldmath{$-0.007$} \\
 & $k = 3$ & 2.5228 & 2.5508 & 2.5226 & 2.5195 & $+0.000$ & \boldmath{$-0.028$} \\
M3ED official & $k = 0$ & 2.4740 & 2.5154 & 2.4740 & 2.6925 & $+0.000$ & \boldmath{$-0.041$} \\
 & $k = 1$ & 1.9487 & 1.9641 & 1.9487 & 1.9665 & $+0.000$ & \boldmath{$-0.015$} \\
 & $k = 2$ & 1.7113 & 1.7147 & 1.7118 & 1.7121 & $-0.001$ & $-0.003$ \\
 & $k = 3$ & 1.5959 & 1.5971 & 1.5968 & 1.5997 & \boldmath{$-0.001$} & $-0.001$ \\
EmoWOZ official & $k = 0$ & 2.0383 & 2.0420 & 2.0383 & 2.0565 & $+0.000$ & $-0.004$ \\
 & $k = 1$ & 1.9573 & 1.9569 & 1.9579 & 1.9636 & $-0.001$ & $+0.000$ \\
 & $k = 2$ & 1.9343 & 1.9325 & 1.9343 & 1.9373 & $-0.000$ & $+0.002$ \\
 & $k = 3$ & 1.9145 & 1.9139 & 1.9145 & 1.9183 & $-0.000$ & $+0.001$ \\
EmotionLines $n = 5$ & $k = 0$ & 4.7824 & 4.8141 & 4.7824 & 4.8269 & $+0.000$ & \boldmath{$-0.032$} \\
 & $k = 1$ & 4.7134 & 4.7360 & 4.7133 & 4.7370 & \boldmath{$+0.000$} & \boldmath{$-0.023$} \\
 & $k = 2$ & 4.6859 & 4.6987 & 4.6845 & 4.7080 & $+0.001$ & \boldmath{$-0.013$} \\
 & $k = 3$ & 4.6755 & 4.6802 & 4.6758 & 4.7178 & $-0.000$ & \boldmath{$-0.005$} \\
 & $k = 4$ & 4.6692 & 4.6703 & 4.6761 & 4.7374 & \boldmath{$-0.007$} & $-0.001$ \\
 & $k = 5$ & 4.7015 & 4.7015 & 4.7015 & 4.7705 & $+0.000$ & $+0.000$ \\
\bottomrule

\end{tabularx}

\end{table}

\section[\appendixname~\thesection]{CCPR-L Decomposition with Confidence~Intervals}
\label{app:ccprl}

Block-bootstrap 95\% CIs (10{,}000 resamples over sessions/series/dialogues) for the three terms introduced in Section~\ref{sec:learned}.
\emph{total} is
CCPR-L$+$head $-$ ccprJ$+$head; \emph{context} is the information set
control CCPR-L$+$head $-$ CCPR-L$_{\mathrm{noctx}}+$head; \emph{table} is
CCPR-L$_{\mathrm{noctx}}+$head $-$ ccprJ$+$head. At~$k = n$ the
completion is empty, so the context term is identically zero and the
control and the treatment coincide by~construction.

\begin{table}[H]
\centering
\small
\caption{CCPR-L decomposition, mean [95\% CI].}
\label{tab:ccprl-ci}

\begin{adjustwidth}{-\extralength}{0cm}
\newcolumntype{C}{>{\centering\arraybackslash}X}
\newcolumntype{B}{>{\raggedright\arraybackslash}X}
\newcolumntype{A}{>{\raggedleft\arraybackslash}X}
\begin{tabularx}{\fulllength}{llCCCc}

\toprule

\textbf{Corpus} & \textbf{Term} & \boldmath{$k = 0$} & \boldmath{$k = 1$} & \boldmath{$k = 2$} & \boldmath{$k = 3$} \\
\midrule
IEMOCAP & total & $-0.0653$ & $-0.0618$ & $-0.0339$ & $+0.0054$ \\
 & & $[-0.1014,-0.0375]$ & $[-0.0892,-0.0400]$ & $[-0.0479,-0.0218]$ & $[+0.0024,+0.0083]$ \\
 & context & $-0.0610$ & $-0.0401$ & $-0.0147$ & $0.0000$ \\
 & & $[-0.0969,-0.0333]$ & $[-0.0616,-0.0219]$ & $[-0.0271,-0.0037]$ & $[0,0]$ \\
 & table & $-0.0043$ & $-0.0218$ & $-0.0192$ & $+0.0054$ \\
 & & $[-0.0072,-0.0013]$ & $[-0.0272,-0.0170]$ & $[-0.0226,-0.0154]$ & $[+0.0024,+0.0083]$ \\
\midrule
M3ED & total & $-0.0946$ & $-0.0360$ & $-0.0138$ & $-0.0030$ \\
official & & $[-0.1170,-0.0687]$ & $[-0.0447,-0.0287]$ & $[-0.0178,-0.0097]$ & $[-0.0113,+0.0035]$ \\
 & context & $-0.0735$ & $-0.0163$ & $-0.0014$ & $0.0000$ \\
 & & $[-0.0998,-0.0459]$ & $[-0.0261,-0.0066]$ & $[-0.0043,+0.0010]$ & $[0,0]$ \\
 & table & $-0.0211$ & $-0.0197$ & $-0.0124$ & $-0.0030$ \\
 & & $[-0.0307,-0.0126]$ & $[-0.0296,-0.0098]$ & $[-0.0175,-0.0074]$ & $[-0.0113,+0.0035]$ \\
\midrule
EmoWOZ & total & $-0.0171$ & $-0.0064$ & $-0.0022$ & $+0.0018$ \\
official & & $[-0.0272,-0.0072]$ & $[-0.0104,-0.0026]$ & $[-0.0062,+0.0018]$ & $[+0.0008,+0.0028]$ \\
 & context & $-0.0146$ & $-0.0073$ & $-0.0038$ & $0.0000$ \\
 & & $[-0.0232,-0.0061]$ & $[-0.0110,-0.0038]$ & $[-0.0071,-0.0005]$ & $[0,0]$ \\
 & table & $-0.0025$ & $+0.0009$ & $+0.0016$ & $+0.0018$ \\
 & & $[-0.0048,-0.0002]$ & $[-0.0004,+0.0022]$ & $[-0.0007,+0.0036]$ & $[+0.0008,+0.0028]$ \\
\bottomrule

\end{tabularx}
\end{adjustwidth}

\end{table}

\section[\appendixname~\thesection]{CCPR-S~Details}

The bias-variance characterization suggests an operator that adapts to
cell density rather than committing to either endpoint. CCPR-S
(Section~\ref{sec:method}) does this with one scalar per level. It stays
within $0.002$ nats of the better endpoint everywhere (only significant
loss: $0.0005$ at IEMOCAP $k = 1$), matches the per-level marginalizer
overall while beating the shared table significantly on IEMOCAP, M3ED and
EmotionLines ($-0.029$/$-0.015$/$-0.012$), and~beats \emph{both}
endpoints at the sparsest cells (M3ED $k = 2/3$; $n = 5,k = 4$:
$-0.007$). Crucially it keeps the shared table serving every $k$ and $n$
and spends its per-level freedom on one completion-reweighting scalar,
not on $(n + 1)$ response tables---the same division of labor as
CCPR-L's $\lambda_k$. Full per-level results:
Appendix~\ref{app:ccprs}.

\section[\appendixname~\thesection]{Mixed Rater-Count~Head-to-Head}

Against the
stronger per-$(n,k)$ \emph{inconsistent marginalizer} (one table per
$(n,k)$, each fit by the same per-level posterior-weighted iteration; oracle-like because it knows
$n$ and trains $n = 5$ tables on rows the shared table never sees), the~shared table wins the sparsest cells ($n = 5,k = 4$: $-0.002$;
$n = 3,k = 2$: $-0.003$) but loses the dense ones, for~an overall cost of
$+0.0085$ nats (CI $[0.005,0.012]$). Consistency is therefore not
predictively superior even against an oracle refit; its value is
zero-configuration service---one table, no per-$(n,k)$ refits---plus
auditability, at~a small, fully quantified~cost.

\section[\appendixname~\thesection]{Experimental Details and~Audit}
\label{app:audit}

The implementation fixes random seeds and uses CPU computations for the
reported table-based estimators. The ancillary archive separates scripts
and stored aggregate summaries from the original licensed data and
per-item outputs. Historical audit scripts are retained as implementation
checks; their hard-coded expected values do not replace independent
verification of the datasets, estimators, or inferential assumptions.
In particular, probability normalization and exact completion
marginalization can be checked separately from predictive performance.
At~$n = 5$, both models violate
$V(k + 1) \le V(k)$ at high $k$ (e.g.,\ $V(4)=4.6693$, $V(5)=4.7087$),
consistent with top-1 overconfidence in the sparse 462-composition space;
a per-level concentration probe never improves test NLL (21 of 22 cells
select the shared $\kappa$; the exception is worse), so this particular scalar concentration grid did not resolve the
observed fitted-risk limitation; it does not establish a unique structural
cause. The same-item, inertia, allocation-baseline, and identity aggregate
summaries accompany the implementation. See the archive README and
manifest for included files, dataset dependencies, and verification limits.

\begin{algorithm}[H]
\caption{CCPR-J: Posterior-weighted moment estimation of the response table}
\label{alg:em}
\begin{algorithmic}[1]
\Require training rows with partial counts $\tilde c_i^{(k)}$, relation $r_i$,
target counts $c_i$; smoothing $s>0$; concentration $\kappa$;
level weights $w_k$; iteration count $M$
\Ensure response table $T$
\State Initialize $T$ as the smoothed mean table over full-panel training rows
\State Fit empirical completion tables $W_k(m\mid\tilde c^{(k)},r)$ on training data
\For{$a=1,\ldots,M$}
  \State Set $\mathrm{num}_g\gets\mathbf{0}$ and $\mathrm{den}_g\gets0$ for every $g$
  \For{$k=0,\ldots,n$}
    \For{each training row $i$}
      \For{each completion $m$ with $|m|=n-k$}
        \State $g_{i,m}\gets g(\tilde c_i^{(k)}+m,r_i)$
        \State $\ell_{i,m}\gets \log\mathrm{DM}(c_i\mid\kappa T_{g_{i,m}})
                     +\log W_k(m\mid\tilde c_i^{(k)},r_i)$
      \EndFor
      \State $p_{i,m}\gets \mathrm{softmax}_m(\ell_{i,m})$
      \For{each completion $m$}
        \State $\mathrm{num}_{g_{i,m}}\gets\mathrm{num}_{g_{i,m}}+w_k p_{i,m}c_i$
        \State $\mathrm{den}_{g_{i,m}}\gets\mathrm{den}_{g_{i,m}}+w_k p_{i,m}$
      \EndFor
    \EndFor
  \EndFor
  \State For each $g$ with $\mathrm{den}_g>0$, set $T_g\gets(\mathrm{num}_g+s\mathbf{1})/(\mathbf{1}^{\top}\mathrm{num}_g+Ks)$; retain $T_g$ otherwise
\EndFor
\end{algorithmic}
\end{algorithm}

\begin{adjustwidth}{-\extralength}{0cm}

\end{adjustwidth}

\begin{thebibliography}{999}

\bibitem[Leonardelli et~al.(2025)Leonardelli, Casola, Peng, Rizzi, Basile, Fersini, Frassinelli, Jang, Pavlovic, Plank, and Poesio]{lewidi2025}
Leonardelli, E.; Casola, S.; Peng, S.; Rizzi, G.; Basile, V.; Fersini, E.; Frassinelli, D.; Jang, H.; Pavlovic, M.; Plank, B.; Poesio, M.
\newblock {LeWiDi-2025} at {NLPerspectives}: Third Edition of the Learning with
  Disagreements Shared Task.
\newblock In \emph{Proceedings of the 4th Workshop on Perspectivist Approaches to NLP}; Association for Computational Linguistics: Suzhou, China, 2025; pp. 182--195.
\newblock \url{https://doi.org/10.18653/v1/2025.nlperspectives-1.16}.

\bibitem[Xu and Jurgens(2026)]{beyondconsensus2026}
Xu, Y.; Jurgens, D.
\newblock Beyond Consensus: Perspectivist Modeling and Evaluation of Annotator
  Disagreement in {NLP}.
\newblock {\em arXiv} {\bf 2026}, arXiv:2601.09065.

\bibitem[Wu et~al.(2024)Wu, Li, Zhang, Chiu, Li, Bai, Sainath, and
  Woodland]{wu2024ambiguity}
Wu, W.; Li, B.; Zhang, C.; Chiu, C.C.; Li, Q.; Bai, J.; Sainath, T.; Woodland,
  P.
\newblock Handling Ambiguity in Emotion: From Out-of-Domain Detection to
  Distribution Estimation.
\newblock In \emph{Proceedings of the 62nd Annual Meeting of the Association for Computational Linguistics (Volume 1: Long Papers)}; Association for Computational Linguistics: Bangkok, Thailand, 2024; pp. 2078--2093.

\bibitem[Wu et~al.(2022)Wu, Zhang, and Woodland]{wu2022disterc}
Wu, W.; Zhang, C.; Woodland, P.C.
\newblock Distribution-Based Emotion Recognition in Conversation.
\newblock {\em arXiv} {\bf 2022}, arXiv:2211.04834.

\bibitem[Li and Chen(2024)]{li2024inldl}
Li, X.; Chen, S.
\newblock No Regularization Is Needed: Efficient and Effective Incomplete Label
  Distribution Learning.
\newblock In \emph{Proceedings of the  Thirty-Third International
  Joint Conference on Artificial Intelligence};  International Joint Conferences on Artificial Intelligence Organization: Jeju, Republic of Korea, 2024; pp. 4470--4478.
\newblock \url{https://doi.org/10.24963/ijcai.2024/494}.

\bibitem[Zhang et~al.(2026)Zhang, Lian, Liu, Takebe, and
  Nakashima]{zhang2026simlabel}
Zhang, L.; Lian, Z.; Liu, H.; Takebe, T.; Nakashima, Y.
\newblock SimLabel: Similarity-Weighted Semi-supervision for Multi-Annotator
  Learning with Missing Labels.
\newblock In \emph{Proceedings of the  AAAI Conference on
  Artificial Intelligence}, AAAI Press: Singapore, 2026; Volume~40, pp. 28328--28336.
\newblock \url{https://doi.org/10.1609/aaai.v40i33.40061}.

\bibitem[Gheibi and Ghazizadeh(2026)]{hardsoft2026}
Gheibi, M.; Ghazizadeh, G.
\newblock Same Target, Different Basins: Hard vs.\ Soft Labels for Annotator
  Distributions.
\newblock {\em arXiv} {\bf 2026}, arXiv:2605.20642.

\bibitem[Kohli(2026)]{saturation2026}
Kohli, G.
\newblock Metric-Dependent Annotation Saturation for Learning from Label
  Distributions.
\newblock {\em arXiv} {\bf 2026}, arXiv:2605.29797.

\bibitem[Sheng et~al.(2008)Sheng, Provost, and Ipeirotis]{sheng2008get}
Sheng, V.S.; Provost, F.; Ipeirotis, P.G.
\newblock Get Another Label? Improving Data Quality and Data Mining Using
  Multiple, Noisy Labelers.
\newblock In \emph{Proceedings of the  14th ACM SIGKDD
  International Conference on Knowledge Discovery and Data Mining}; Association for Computing Machinery: Las Vegas, NV, USA, 2008; pp.
  614--622.
\newblock {\url{https://doi.org/10.1145/1401890.1401965}}.

\bibitem[Hsueh et~al.(2009)Hsueh, Melville, and
  Sindhwani]{hsueh2009dataquality}
Hsueh, P.Y.; Melville, P.; Sindhwani, V.
\newblock Data Quality from Crowdsourcing: A Study of Annotation Selection
  Criteria.
\newblock In \emph{Proceedings of the  {NAACL} {HLT} 2009 Workshop
  on Active Learning for Natural Language Processing}; Association for Computational Linguistics: Boulder, CO, USA,   2009; pp. 27--35.

\bibitem[Yan et~al.(2011)Yan, Rosales, Fung, and Dy]{yan2011activecrowds}
Yan, Y.; Rosales, R.; Fung, G.; Dy, J.G.
\newblock Active Learning from Crowds.
\newblock In \emph{Proceedings of the  28th International
  Conference on Machine Learning}; Omnipress: Bellevue, WA, USA, 2011.

\bibitem[Li et~al.(2015)Li, Jiang, and Zhou]{li2015multilabelcrowds}
Li, S.Y.; Jiang, Y.; Zhou, Z.H.
\newblock Multi-Label Active Learning from Crowds.
\newblock {\em arXiv} {\bf 2015}, arXiv:1508.00722.

\bibitem[Wang and Plank(2023)]{wang2023actor}
Wang, X.; Plank, B.
\newblock {ACTOR}: Active Learning with Annotator-specific Classification Heads
  to Embrace Human Label Variation.
\newblock In \emph{Proceedings of the  2023 Conference on Empirical
  Methods in Natural Language Processing};  Association for Computational
  Linguistics: Singapore,  2023; pp. 2046--2052.
\newblock {\url{https://doi.org/10.18653/v1/2023.emnlp-main.126}}.

\bibitem[van~der Meer et~al.(2024)van~der Meer, Falk, Murukannaiah, and
  Liscio]{vandermeer2024acal}
van~der Meer, M.; Falk, N.; Murukannaiah, P.K.; Liscio, E.
\newblock Annotator-Centric Active Learning for Subjective {NLP} Tasks.
\newblock In \emph{Proceedings of the  2024 Conference on Empirical
  Methods in Natural Language Processing}; Association for Computational
  Linguistics: Miami, FL, USA, 2024; pp. 18537--18555.
\newblock {\url{https://doi.org/10.18653/v1/2024.emnlp-main.1031}}.

\bibitem[Li et~al.(2021)Li, Zhu, Li, Wang, Li, Liao, and Zheng]{li2021emotion}
Li, D.; Zhu, X.; Li, Y.; Wang, S.; Li, D.; Liao, J.; Zheng, J.
\newblock Emotion Inference in Multi-Turn Conversations with Addressee-Aware
  Module and Ensemble Strategy.
\newblock In \emph{Proceedings of the  2021 Conference on Empirical
  Methods in Natural Language Processing}; Association for Computational
  Linguistics: Punta Cana, Dominican Republic,   2021; pp. 3935--3941.

\bibitem[Koga et~al.(2024)Koga, Kando, and Miyao]{koga2024forecasting}
Koga, Y.; Kando, S.; Miyao, Y.
\newblock Forecasting Implicit Emotions Elicited in Conversations.
\newblock In \emph{Proceedings of the  17th International Natural
  Language Generation Conference}; Association for Computational Linguistics: Tokyo, Japan,  2024; pp. 145--152.

\bibitem[Wang et~al.(2026)Wang, Mai, Tao, Lin, Tong, Pan, Yan, Wang, and
  Gao]{wang2026hief}
Wang, H.; Mai, X.; Tao, Z.; Lin, J.; Tong, X.; Pan, I.; Yan, S.; Wang, Y.; Gao,
  S.
\newblock Hi-EF: Benchmarking Emotion Forecasting in Human-interaction.
\newblock In \emph{Proceedings of the  AAAI Conference on
  Artificial Intelligence}; AAAI Press: Singapore, 2026; Volume~40, pp. 2110--2118.
\newblock \url{https://doi.org/10.1609/aaai.v40i3.37193}.

\bibitem[Noor et~al.(2026)Noor, Latif, Shahzad, and Fatima]{noor2026affect}
Noor, S.; Latif, S.; Shahzad, R.K.; Fatima, M.
\newblock From Affect Prediction to Affect Forecasting: Evidence for Distinct
  Information Sources in Longitudinal Text.
\newblock {\em arXiv} {\bf 2026}, arXiv:2606.16687.

\bibitem[Inoshita and Ueno(2026)]{inoshita2026bsetd}
Inoshita, K.; Ueno, T.
\newblock Bayesian Spectral Emotion Transition Discovery from Multi-Annotator
  Disagreement.
\newblock {\em arXiv} {\bf 2026}, arXiv:2606.01906. 

\bibitem[Dickey et~al.(1987)Dickey, Jiang, and Kadane]{dickey1987censored}
Dickey, J.M.; Jiang, J.M.; Kadane, J.B.
\newblock Bayesian Methods for Censored Categorical Data.
\newblock {\em J. Am. Stat. Assoc.} {\bf 1987},
  {\em 82},~773--781.

\bibitem[Fox(1966)]{fox1966marginal}
Fox, B.
\newblock Discrete Optimization via Marginal Analysis.
\newblock {\em Manag. Sci.} {\bf 1966}, {\em 13},~210--216.

\bibitem[Busso et~al.(2008)Busso, Bulut, Lee, Kazemzadeh, Mower, Kim, Chang,
  Lee, and Narayanan]{busso2008iemocap}
Busso, C.; Bulut, M.; Lee, C.C.; Kazemzadeh, A.; Mower, E.; Kim, S.; Chang,
  J.N.; Lee, S.; Narayanan, S.S.
\newblock IEMOCAP: Interactive Emotional Dyadic Motion Capture Database.
\newblock {\em Lang. Resour. Eval.} {\bf 2008}, {\em
  42},~335--359.

\bibitem[Zhao et~al.(2022)Zhao, Zhang, Hu, Liu, Jin, Wang, and
  Li]{zhao2022m3ed}
Zhao, J.; Zhang, T.; Hu, J.; Liu, Y.; Jin, Q.; Wang, X.; Li, H.
\newblock M3ED: Multi-modal Multi-scene Multi-label Emotional Dialogue
  Database.
\newblock In \emph{Proceedings of the  60th Annual Meeting of the
  Association for Computational Linguistics (Volume 1: Long Papers)}; Association for Computational Linguistics: Dublin, Ireland, 2022; pp. 5699--5710.

\bibitem[Feng et~al.(2022)Feng, Lubis, Geishauser, Lin, Heck, van Niekerk, and
  Ga{\v{s}}i{\'c}]{feng2022emowoz}
Feng, S.; Lubis, N.; Geishauser, C.; Lin, H.C.; Heck, M.; van Niekerk, C.;
  Ga{\v{s}}i{\'c}, M.
\newblock EmoWOZ: A Large-Scale Corpus and Labelling Scheme for Emotion
  Recognition in Task-Oriented Dialogue Systems.
\newblock In \emph{Proceedings of the  Thirteenth Language
  Resources and Evaluation Conference}; European Language Resources Association: Marseille, France, 2022; pp.
  4096--4113.

\bibitem[Chen et~al.(2018)Chen, Hsu, Kuo, Huang, and Ku]{chen2018emotionlines}
Chen, S.Y.; Hsu, C.C.; Kuo, C.C.; Huang, T.H.; Ku, L.W.
\newblock EmotionLines: An Emotion Corpus of Multi-Party Conversations.
\newblock In \emph{Proceedings of the Eleventh International Conference on Language Resources and Evaluation (LREC 2018)}; European Language Resources Association: Miyazaki, Japan, 2018; pp. 1597--1601.
\newblock \url{https://aclanthology.org/L18-1252/}.

\end{thebibliography}
\end{document}